\documentclass[runningheads]{llncs}

\usepackage{eccv}

\usepackage{eccvabbrv}

\usepackage{graphicx}
\usepackage{booktabs}
\usepackage{multirow}
\usepackage{amssymb}
\usepackage{pifont}
\usepackage{tabularx}
\usepackage{caption}
\usepackage{subcaption}

\usepackage[accsupp]{axessibility}  

\usepackage[pagebackref,breaklinks,colorlinks,citecolor=eccvblue]{hyperref}

\usepackage{orcidlink}
\usepackage{array}

\newcommand\donghwa[1]{\textcolor{orange}{#1}} 

\newcommand{\cmark}{\textcolor{green}{\ding{51}}} 
\newcommand{\xmark}{\textcolor{red}{\ding{55}}}   

\begin{document}

\title{Finding NeMo: Negative-mined Mosaic Augmentation for Referring Image Segmentation} 

\titlerunning{Finding NeMo: Negative-mined Mosaic Augmentation}

\author{Seongsu Ha\inst{1,2}\thanks{Equal contribution}\protect\footnote[4]{Work done while at Seoul National University.}
\orcidlink{0009-0007-3035-2591}
\and
Chaeyun Kim\inst{1}\protect\footnotemark[1]\orcidlink{0009-0006-6661-4437}
\and
Donghwa Kim\inst{1}\protect\footnotemark[1]\orcidlink{0009-0002-7931-5648}
\and \\
Junho Lee\inst{1}\protect\orcidlink{0000-0001-7643-1024}
\and
Sangho Lee\inst{3}\protect\orcidlink{0009-0004-8718-0004}
\and
Joonseok Lee\inst{1,4}\thanks{Corresponding author}\orcidlink{0000-0002-0786-8086}
}

\authorrunning{S.~Ha, C.~Kim, D.~Kim et al.}

\institute{Seoul National University, Seoul, Korea
\and
Twelve Labs, Seoul, Korea \\
\and
Allen Institute for AI, Seattle, Washington, USA \\
\and
Google Research, Mountain View, California, USA \\
\email{
mars@twelvelabs.io, \{golddohyun,ehd9712,joon2003\}@snu.ac.kr,\\
 sanghol@allenai.org, joonseok@snu.ac.kr}}

\maketitle


\begin{abstract}
Referring Image Segmentation is a comprehensive task to segment an object referred by a textual query from an image. In nature, the level of difficulty in this task is affected by the existence of similar objects and the complexity of the referring expression. Recent RIS models still show a significant performance gap between easy and hard scenarios. We pose that the bottleneck exists in the data, and propose a simple but powerful data augmentation method, Negative-mined Mosaic Augmentation (NeMo). This method augments a training image into a mosaic with three other negative images carefully curated by a pretrained multimodal alignment model, \emph{e.g.}, CLIP, to make the sample more challenging. We discover that it is critical to properly adjust the difficulty level, neither too ambiguous nor too trivial. The augmented training data encourages the RIS model to recognize subtle differences and relationships between similar visual entities and to concretely understand the whole expression to locate the right target better. Our approach shows consistent improvements on various datasets and models, verified by extensive experiments.
\end{abstract}

\section{Introduction}
\label{sec:intro}

\begin{figure}[t]
    \centering
    \includegraphics[width=0.5\linewidth]{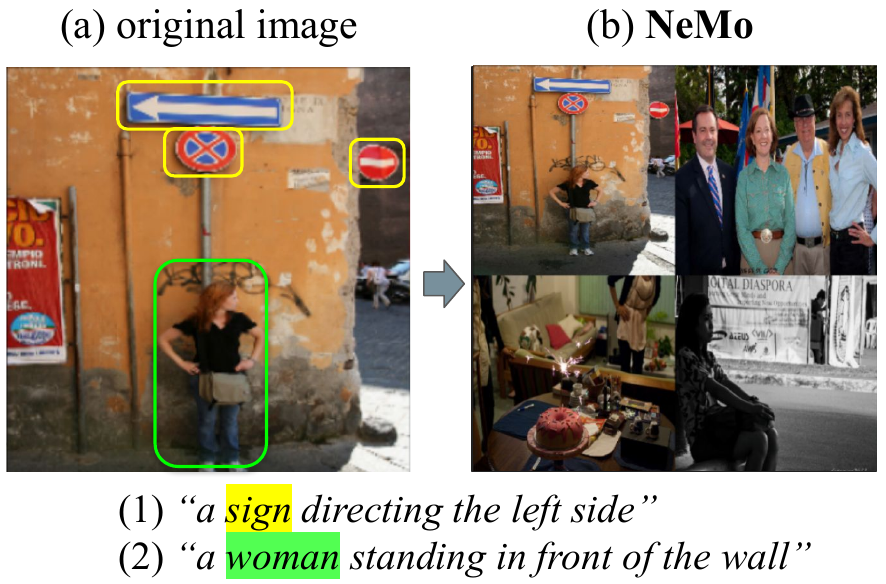}
    \caption{Diverse visual and linguistic challenges of referring scenarios. In (a), query (1) demands discernment among three road signs, while query (2) involves identifying a ``woman'', relatively easier due to a single instance. NeMo, our method, in (b) uses similar negative images to generate a mosaic. Query (2) becomes harder as the augmented image contains additional instances of ``woman'' (\emph{e.g.}, women standing or sitting), and thus ``in front of the wall'' becomes crucial hint to solve the problem.}
    \label{fig:vis_text_diff_example}
\end{figure}

Referring Image Segmentation (RIS) is a fundamental task in computer vision that aims to segment objects described in a natural language expression within a given scene.
Central to RIS is not merely the visual recognition of objects but also the intricate understanding of the interrelationships among these objects, interpreted through linguistic cues. 

Each RIS problem often requires a different level of multimodal understanding capabilities, depending on visual ambiguity as well as linguistic complexity.
For instance, having visually similar objects to the referent in an image complicates locating and identifying the correct object. In such a case, precise comprehension of the referring expression becomes a key to find the right target.

\begin{figure}[htbp]
  \centering
  \begin{minipage}{.5\textwidth}
    \captionof{table}{Statistics of representative Referring Image Segmentation (RIS) Datasets}
    \label{tab:ris_dataset_stats}
    \footnotesize
    \resizebox{0.98\linewidth}{!}{
    \begin{tabular}{lccc}
      \toprule
      Dataset             & RefCOCO & RefCOCO+ & G-Ref \\ \midrule
      \# Images           & 19,994  & 19,992   & 26,711 \\
      \# Ref. Exp. & 142,209 & 141,564  & 85,474 \\ 
      Query length   & 3.61    & 3.53     & 8.43   \\ 
      Obj. per query & 1.76    & 1.67     & 3.03   \\
      \bottomrule
    \end{tabular}}
  \end{minipage}%
  \hspace{0.1cm}
  \begin{minipage}{.46\textwidth}
    \centering
    \captionof{table}{mIoU \& oIoU on 100 easy and hard samples from G-Ref UMD test set}
    \label{tab:intro_easy_hard_sample}
    \footnotesize
    \renewcommand{\tabcolsep}{4pt}
    \resizebox{0.99\linewidth}{!}{
    \begin{tabular}{lcccc}
      \toprule
      & \multicolumn{2}{c}{\textbf{mIoU}} & \multicolumn{2}{c}{\textbf{oIoU}} \\ 
      \cmidrule(lr){2-3} \cmidrule(lr){4-5}
      \textbf{Models} & \textbf{Easy} & \textbf{Hard} & \textbf{Easy} & \textbf{Hard} \\ 
      \midrule
      LAVT~\cite{yang2022lavt} & 78.26 & 54.61 & 79.16 & 47.40 \\
      CRIS~\cite{wang2022cris} & 76.89 & 52.97 & 78.81 & 43.20 \\
      CGFormer~\cite{tang2023cgformer} & 79.86 & 61.22 & 79.95 & 53.27 \\
      \bottomrule
    \end{tabular}}
  \end{minipage}
\end{figure}

\begin{figure}
\centering
\includegraphics[width=0.8\linewidth]{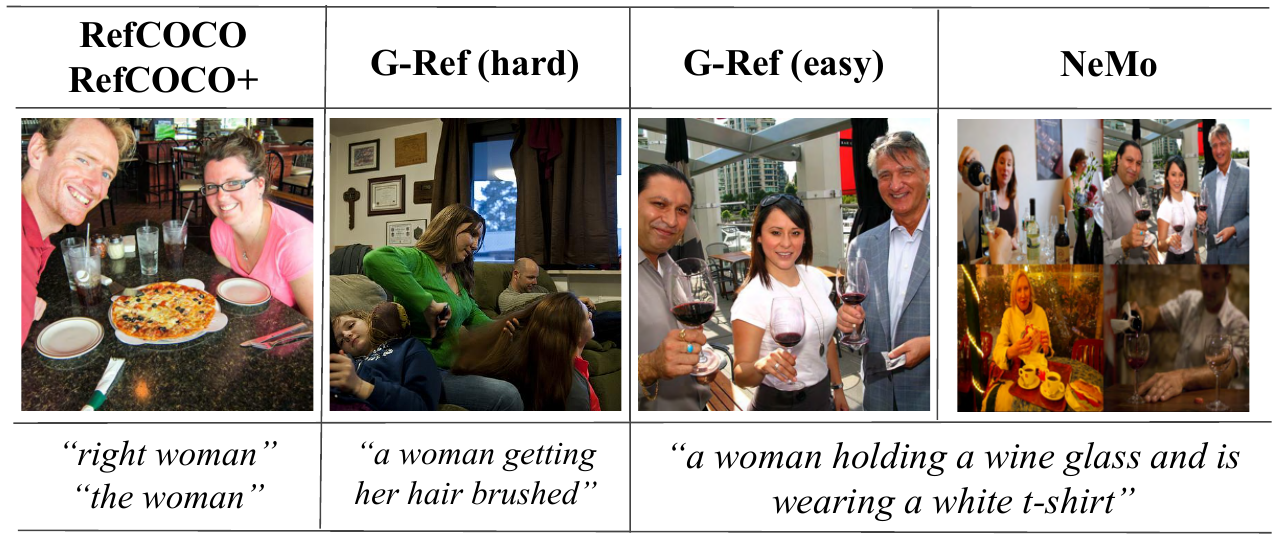} 
\caption{Data samples from RIS benchmarks and augmented samples using our NeMo. RefCOCO and RefCOCO+ are characterized by relatively easier scenarios with simple referring expressions, whereas G-Refs encompass more challenging sets.}
\label{fig:dataset_sample} 
\end{figure}

\cref{fig:vis_text_diff_example}(a) illustrates varying degrees of difficulties even within the same image. 
The sentence (1) is relatively harder, since three road signs exist and the model needs to find the target by precisely understanding the entire phrase.
The expression (2), on the other hand, is relatively easier, since there is only one woman, so one can easily find her regardless of the rest of the phrase.

Many existing RIS datasets, however, have not been created considering such challenge levels; rather, many examples can be solved by simply finding an object corresponding to the referred class.
RefCOCO and RefCOCO+~\cite{yu2016modeling}, for example, contain easier scenarios with less visual ambiguity and linguistic complexity.
On the contrary, G-Ref~\cite{mao2016generation} is considered harder.
As in \cref{tab:ris_dataset_stats} and \cref{fig:dataset_sample}, it contains more objects in each image on average, and queries are relatively longer. 

The level of difficulty significantly varies even within the same dataset.
To illustrate, we manually select 100 easy and hard samples from the G-Ref UMD test set.
We select `easy' samples containing only a single object per its category, straightforward to identify the target without ambiguity.
For instance, in the third image in \cref{fig:dataset_sample}, there is only a single woman holding a glass, so just finding `a woman' will suffice.
Conversely, `hard' examples contain multiple objects within the same category, necessitating a detailed perception to distinguish the intended target.
Specific easy and hard examples are provided in \cref{sec:easy_hard_examples}.

\cref{tab:intro_easy_hard_sample} shows a significant performance gap between the easy and hard samples by recent RIS models.
This indicates that they are capable of picking the right object without ambiguity, but they tend to lack in understanding delicate meaning in the referring expressions and using them to distinguish multiple objects in the same (or similar) categories.
To further improve the RIS performance, this observation reveals that we may need to revisit if the models have been provided with sufficiently difficult training data to learn from.

Given this problem landscape, we aim to improve the performance by tackling the data part of the training.
Specifically, we postulate that amplifying the exposure of the models to challenging examples at training could fortify their capability to understand the subtle dynamics between visual and linguistic components.
Such complexity often arises when multiple objects, potentially of the same class, coexist within an image, encouraging the model to fully understand both the scene and the given referring expression.
However, manually labeling such `hard' data examples is prohibitively expensive.

Recognizing the key factors behind the difficulty and quality of RIS training data, we introduce a simple but universally applicable data augmentation method, Negative-mined Mosaic Augmentation (NeMo).
Inspired by the mosaic augmentation in YOLO v4~\cite{bochkovskiy2020yolov4}, NeMo augments each training image by combining it with three other images in a $2 \times 2$ formation, showing four times more objects on average.
However, NeMo differs from the previous method in that the extra three images are not chosen at random, but are carefully selected to create a properly challenging training example.
Specifically, we propose considering relevance between the referring expression and candidate images, measured by a cross-modal retrieval model, \emph{e.g.}, CLIP.
To build a mosaic, our method selects negative images containing objects from the same or similar category to the referred object, retrieved based on the relevance to the referring expression.
Augmented mosaic images mimic challenging referring examples as in \cref{fig:vis_text_diff_example}(b), encouraging the model to learn subtle visual differences and to concretely understand the given referring expression to better locate the target.


One might concern if combining similar images may create false positives, where the correct object in an image becomes invalid due to the objects in the other quadrants.
We study the possibility and impact of such false positives, and discover that it is indeed critical
for the mosaic to have the right level of difficulty to be maximally effective.
Based on our observations, we present a strategic retrieval process to make the mosaic neither too hard nor too easy.

From extensive experiments with five state-of-the-art RIS models, we verify that NeMo consistently improves performance across all models on multiple datasets.
Furthermore, we exhibit that NeMo encourages a model to make better connections between words and visual components, grasping fine details in the scene and the referring expression. 
We expect our study to support the primary aim of the RIS task, distinguishing multiple candidate objects in the scene and recognizing the target based on the textual description.


Our main contributions are summarized as follows:
\begin{enumerate}
  \item We introduce NeMo, a simple but powerful labor-free data augmentation method for Referring Image Segmentation (RIS), effective across various datasets and models.
  \item We discover that it is critical to adjust the level of difficulty to successfully apply a mosaic augmentation, and propose a systematic way to tune this difficulty by generating training examples at a properly controlled difficulty.
  \item We empirically verify that NeMo enhances both visual and textual understanding capabilities for segmenting the right target.
\end{enumerate}
\section{Related Work}
\label{sec:related}

\textbf{Referring Image Segmentation (RIS).} Existing studies~\cite{hu2016segmentation} have concentrated on encoder and decoder architectures to handle multimodal features: RNNs~\cite{li2018referring,liu2017recurrent,margffoy2018dynamic,shi2018key} and Transformers~\cite{bellver2020refvos} for text features, and CNNs~\cite{hu2016segmentation,liu2017recurrent}, DeeplabV3~\cite{bellver2020refvos,chen2017rethinking,li2018referring}, and DarkNet~\cite{jing2021locate,luo2020multi} for visual encoding.
Transformer-based backbones~\cite{jing2021locate,liu2021swin,devlin2018bert,kamath2021mdetr} and multi-scale features~\cite{chen2019see,hu2020bi,hui2020linguistic,ye2019cross} are recently popular to capture detailed object masks.
Visual-linguistic fusion has evolved from simple concatenation~\cite{hu2016segmentation} to syntax-based~\cite{huang2020referring,hui2020linguistic,yu2018mattnet} and attention-based; to name some:
LAVT~\cite{yang2022lavt}, VPD~\cite{zhao2023unleashing}, and RefSegformer~\cite{wu2022towards}.
VLT~\cite{ding2021vision}, CRIS~\cite{wang2022cris}, and ReSTR~\cite{kim2022restr} employ cross-modal decoder fusion. ReLA~\cite{liu2023gres} and 
CGformer~\cite{tang2023cgformer} organizes visual features into language-conditioned tokens, capturing region-level information. PolyFormer~\cite{liu2023polyformer} converts grounding tasks into sequential polygon generation.
VPD~\cite{zhao2023unleashing} leverages a multi-scale feature map from a text-to-image diffusion model for RIS.
Unlike prior studies focusing on architectural enhancements,
our work redirects attention to the quality and nature of the data, proposing a straightforward augmentation to create more refined training examples.

\vspace{0.1cm} 
\noindent
\textbf{Datasets for RIS.}
Initially, ReferIt~\cite{kazemzadeh2014referitgame}, RefCOCO~\cite{yu2016modeling} and RefCOCO+~\cite{yu2016modeling} have been introduced as benchmarks for single-target RIS.
RefCOCO contains many positional words such as ``front'' or ``the third from the right'', while RefCOCO+ prohibits such direct usage of words on positions. 
Many examples in these benchmarks often present over-simplified scenarios with a short query, without enough ambiguities in images.
In contrast, recent datasets embrace more complex scenes and intricate linguistic expressions.
G-Ref~\cite{mao2016generation} contains relatively longer sentences, making the task more challenging.
Built upon \cite{krishna2016visual}, PhraseCut~\cite{wu2020phrasecut} provides examples with multiple targets with attributes, categories, and relationships in phrases.
GRES~\cite{liu2023gres} steps up the complexity by incorporating references to no target and multiple objects.
CGFormer~\cite{tang2023cgformer} introduces new splits on RefCOCO datasets to measure generalization capability on objects of unseen classes.
Our method is aligned with these studies for better data quality, but we directly generate complex examples via augmentation.

\vspace{0.1cm} 
\noindent
\textbf{Data Augmentation Methods.}
Various data augmentation methods have been proposed for semantic segmentation.
Early strategies involve random object pasting~\cite{ghiasi2021simple}.
Multi-modal mixup~\cite{hao2023mixgen,zhang2017mixup} fuses images and associated texts.
Beyond this, CutMix~\cite{yun2019cutmix} transposes rectangular image sections onto others. Mosaic data augmentation, originally pioneered for object detection by YOLO v4 \cite{bochkovskiy2020yolov4}, generates a composite image from segments of multiple sources, preserving their ground truths.
MixGen~\cite{hao2023mixgen} is another multimodal augmentation, blending two images and combining pairs of text sequences.
However, this mosaic-based method has not been specifically designed for the RIS task, to the best of our knowledge.
In this paper, we discover that simply putting four images into a mosaic does not guarantee maximal performance improvement, and propose a way specific to RIS to curate mosaics.

\vspace{0.1cm} 
\noindent
\textbf{Retrieval Augmented Vision-Language Models.}
Retrieval Augmented models~\cite{chen2022murag} employ retrieval to leverage additional knowledge from external data to enhance model's learning capabilities.
Starting from KNN-LM~\cite{khandelwal2019generalization}, several works in natural language processing have enhanced large language models by connecting them with external sources, intricately arrayed in structured syntax and semantic relations~\cite{borgeaud2022improving, guu2020retrieval, lewis2020retrieval, liu2020k, peters2019knowledge, yu2021dict}.
This approach has expanded to various vision-language tasks; \emph{e.g.}, image synthesis~\cite{blattmann2022retrieval, sheynin2022knn, chen2022re}, classification~\cite{long2022retrieval}, and multi-modal applications~\cite{yasunaga2023retrieval}.
This method is used to generate hard negatives in tasks like multimedia event extraction~\cite{li2022clip} and provide task-specific data augmentation~\cite{long2022retrieval, liu2023learning}.
Additionally, RA-CLIP~\cite{xie2023ra}, K-Lite~\cite{shen2022k}, and ASIF~\cite{norelli2022asif} employ retrieval to enrich visual concepts and to align image-text modalities.

In our work, we employ retrieval for mosaic augmentation tailored for the RIS task.
By generating ambiguous scenarios with hard negatives, we enforce the model to better connect textual expressions with visual components, aiming to improve the model performance in complex situations.



\begin{figure}[t]
  \centering
  \includegraphics[width=0.85\linewidth]{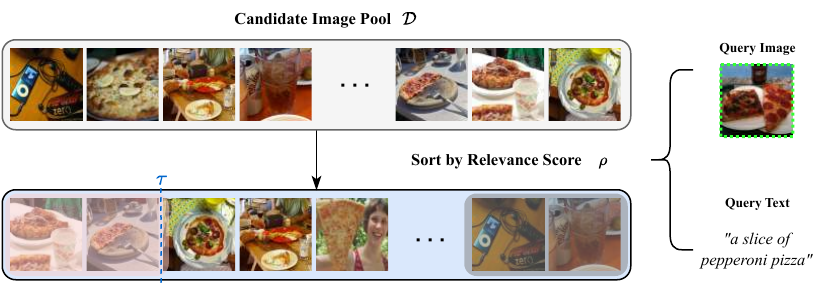} 
  ~
  \includegraphics[width=0.51\linewidth]{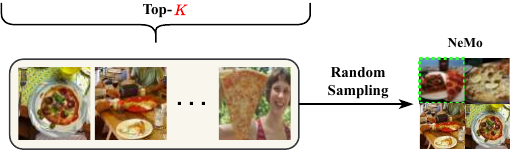} 
  \caption{\textbf{Overall NeMo pipeline.} Given an image and a query, it selects negative images at a proper level of difficulty, filtering out visually or semantically images to the query to avoid false negatives and irrelevant (easy) images identified by text-to-image retrieval. It randomly selects three among the remaining to construct a mosaic.}
  \label{fig:overall_arch}
\end{figure}

\begin{figure*}
  \centering
  \includegraphics[width=\linewidth]{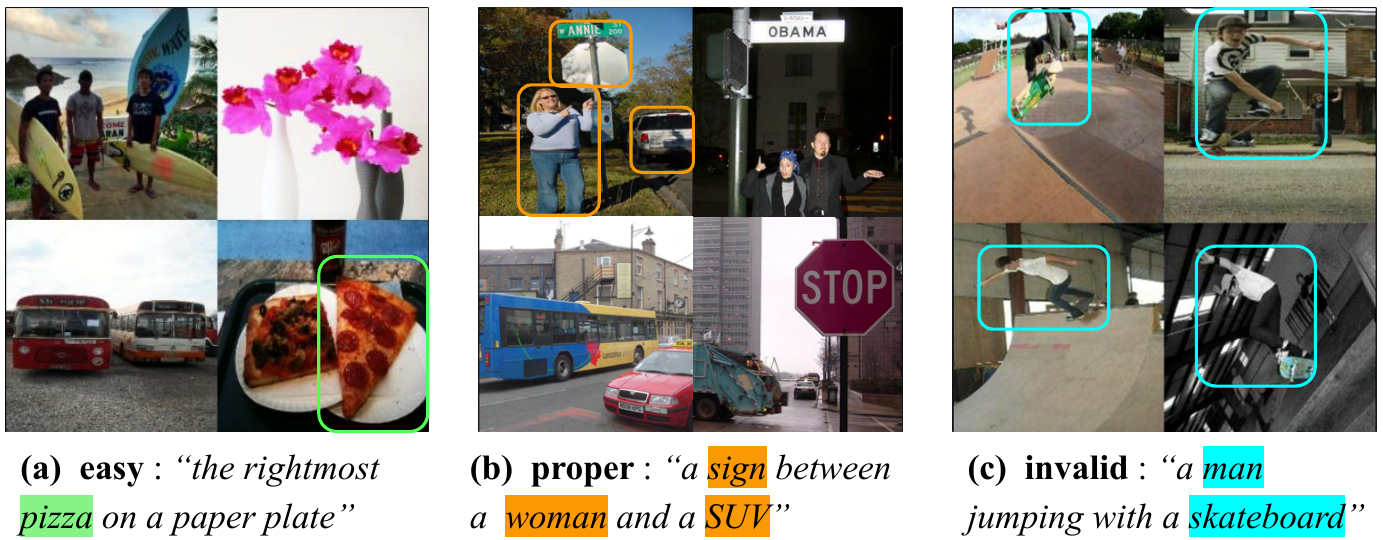}
  \caption{\textbf{Comparison of negative image choices} Finding the ``rightmost pizza'' in (a) is nearly as \textit{easy} as in the single image, as there is no other pizza-like object. Multiple road signs in (b) require discerning the relative location of a woman and an SUV, more \textit{challenging} than the original single image. (c) is \textit{invalid} as multiple images contain ``a man jumping with a skateboard''.}  
  \label{fig:selecting_negatives}
\end{figure*}

\section{NeMo: Negative-mined Mosaic Augmentation}
\label{sec:method}
In this section, we propose a model-agnostic data augmentation framework, Negative-mined Mosaic Augmentation (NeMo), for the RIS task.
Particularly, we introduce a simple but flexible augmentation technique that guides us in choosing proper negative images to create harder training samples.
\cref{fig:overall_arch} illustrates the overall pipeline of our approach.

\subsection{Motivation for Negative Mining}
\label{sec:method:nemo}


As discussed in \cref{sec:intro}, the level of ambiguity in images largely depends on whether they contain objects of similar categories related to the referring expression.
We leverage this nature of the RIS task to control the difficulty of each problem instance by carefully mining the negatives in the mosaic.
\cref{fig:selecting_negatives} demonstrates how the task difficulty varies with the visual and contextual relationships between the negative images and the text query.
For example, a distinct positive image of pizza combined with unrelated images like flowers or buses in (a) poses little challenge for the model.
Conversely, images of multiple skateboarders jumping in (c) provide too many plausible choices for the given phrase, leading to confusion.
We aim to craft training samples with the right level of difficulty as in (b), neither too easy nor too hard. Balancing challenge and distinguishability can foster fine-grained learning of the relationships between the referring expression and the objects in the scene.

To select suitable negative images, we define two hyperparameters as illustrated in \cref{fig:overall_arch}: uni-modal or cross-modal similarity score threshold ($\tau$) to exclude overly similar examples, and the number of top negative image candidates ($K$) to consider.
The filtering process is detailed below.

\subsection{Negative Image Mining Methodology}
\label{sec:neg_img_selection}

For an original training example $(I, T)$, where $I$ and $T$ stand for the original image and referring expression, our approach aims to retrieve negative images that are moderately aligned with $T$ from the pool of all images, $\mathcal{D}$.
To quantify the relevance of each candidate image $I^{(i)} \in \mathcal{D}$, we rely on visual-text similarity scores estimated by a pre-trained cross-modal model, \emph{e.g.}, CLIP~\cite{li2022clip}.

\noindent
\textbf{Determining the Upper-bound.}
At a glance, it looks straightforward to choose the top $N$ images that are most relevant to the target text $T$ as hard negatives.
The images with the highest relevance in $\mathcal{D}$, however, can be visually too similar to the query, which may result abundant potential choices.
Those negatives are in fact `false negatives', where there exists another object perfectly in accordance with the referred expression $T$ within the retrieved negative image and thus
it becomes impossible to find the intended one, as in \cref{fig:selecting_negatives}(c).

To address this, we filter out potential false negatives, or excessively relevant candidate images to $T$.
Specifically, we compute the relevance score $\rho^{(i)}$ of each candidate image $I^{(i)} \in \mathcal{D}$ with the referring expression $T$ by ${\mathbf{t}}^\top \mathbf{v}^{(i)}$, 
where $\mathbf{v}^{(i)}$ and $\mathbf{t}$ are CLIP visual and text embedding of the candidate image $I^{(i)}$ and $T$, respectively.
NeMo then prevents potential false negatives by excluding candidate images that are too relevant to $T$ with $\rho^{(i)} \ge \tau$, where 
$\tau$ is a hyperparameter to control the tolerance of upper-bound filtering.
With a large $\tau$, we filter only extremely relevant images from the candidate set, while with a smaller $\tau$, we aim to filter more to ensure less false negatives to occur.

Alternatively, we may use $\rho$ for the image-to-image similarities, $\mathbf{v}^\top \mathbf{v}^{(i)}$, between the positive image $I$ and all other negative candidates $I^{(i)} \in \mathcal{D}$, where $\mathbf{v}$ is the CLIP visual embedding of $I$.
This can be useful to capture a highly relevant candidate image captioned with a less similar text form.
It is also possible to use both text-to-image (t-i) and image-to-image (i-i) similarities, filtering out images either $\rho_\text{t-i}^{(i)} \ge \tau_\text{t-i}$ or $\rho_\text{i-i}^{(i)} \ge \tau_\text{i-i}$.
See~\cref{sec:sim_score_opt} for empirical comparison.

\vspace{0.1cm} 
\noindent
\textbf{Determining the Lower-bound.}
After we filter out excessively similar images that are $\rho^{(i)} \ge \tau$, we collect $K$ most plausible images described by $T$ among the remaining candidate images.
At this step, the relevance of an image $I^{(i)}$ can be computed using the same $\rho$ or some other $\rho'$.
To keep the framework general, we use $\rho'$ for the relevance used in this step; that is, we collect the top $K$ images with highest $\rho'$ from the set $\{I^{(i)} \in \mathcal{D} : \rho^{(i)} < \tau \}$.
We illustrate the simpler case with $\rho = \rho'$ in \cref{fig:overall_arch}, and the general case with $\rho \ne \rho'$ in~\cref{sec:sim_score_opt}. We then randomly select 3 out of the $K$ candidates, arrange them with the positive $I$ in a $2 \times 2$ mosaic grid, and resize the resulting image to half width and height.
The quadrant corresponding to the positive image is labeled accordingly, while the other three quadrants are set to negative for all pixels.


$K$ is another hyperparameter to control the difficulty of an augmented image.
Lower $K$ would select more plausible images related to $T$, which might include other partially correct objects.
With a higher $K$, chances of choosing less relevant images increase.
When $K \approx |\mathcal{D}|$, it is equivalent to the uniform selection.

\vspace{0.1cm} 
\noindent
\textbf{Augmentation Ratio $\gamma$.}
In practice, we expose the model to single images as well as augmented ones during training, since eventually the model performs on single images.
Specifically, we apply NeMo with a probability of $\gamma \in [0, 1]$, while the rest ($1 - \gamma$) uses an original single image.

\noindent
\textbf{Summary.}
The overall process ensures a proper level of difficulty, guided by the parameters $\tau$ and $K$.
When adjusted properly, $\tau$ helps to filter out images that are too similar, and $K$ determines the number of relevant images to consider.
Careful calibration of them guides the chosen images to be sufficiently similar to $T$ but at the same time distinct enough from the positive image $I$.
Such a balanced selection of images helps the model to better understand and adapt to various visual contexts, thereby significantly improving its learning capabilities.
\subsection{Addressing False Negatives \& False Positives} 
\noindent
Even with a careful choice of $\tau$ and $K$, NeMo may still generate false negatives (FN) and false positives (FP), especially on simpler referring expressions. 

First, FN occurs when another object perfectly matching with $T$ exists in one of the chosen negative images.
This seems problematic since it is still labeled negative but it is essentially a positive.
Nevertheless, we claim that FNs are not significantly detrimental to performance. 
Similar to the Masked Language Modeling in BERT~\cite{devlin2018bert}, where multiple valid fill-ins exist for a blank but only one is taught, the model would learn the probability distribution of plausible objects upon sufficient examples and repeated training.



Meanwhile, FP happens when the correct object in $I$ is affected by its placement in relation to other images, often due to positional descriptors\footnote{\emph{e.g.}, left, right, low, high, top, bottom, o'clock, corner} in $T$. This is more common when the target object is designated by its relationship with other objects or its position within the image frame.
For example, if ``a woman in the left'' is positioned in the right quadrant within the $2 \times 2$ grid and another woman appears in the left negative image, the target designation shifts, invalidating the label on the original target.

FPs can be a more critical problem than FNs, since they can \textit{mislead} the model to select a wrong object.
We show that our NeMo inherently provides a way to prevent FPs; choosing a lower $\tau$ would reduce the chance of FP by filtering out even less relevant candidate images.
See \cref{sec:exp:analysis} for empirical verification. \\





\section{Experiments}
\label{sec:exp}

\subsection{Experimental Settings} 
\label{sec:exp:setting}

\begin{table*}[t]
\centering
\caption{
Overall RIS performance (in oIoU) comparison with and without NeMo}
\label{tab:performance_metrics}
\small
\renewcommand{\tabcolsep}{5pt} 
\resizebox{0.99\linewidth}{!}{
\begin{tabular}{lc|ccc|ccc|cc|c|c}
\toprule
\multirow{2}{*}{RIS model} & \multirow{2}{*}{NeMo} & \multicolumn{3}{c|}{RefCOCO (UNC)} & \multicolumn{3}{c|}{RefCOCO+ (UNC)} & \multicolumn{2}{c|}{G-Ref (UMD)} & GRES & \multirow{2}{*}{Average Gain}  \\
& & Val & TestA & TestB & Val & TestA & TestB & Val & Test & Val & \\
\midrule
\multirow{2}{*}{LAVT~\cite{yang2022lavt}} & \xmark & 72.73 & 75.82 & 68.79 & 62.14 & 68.38 & 55.10 & 61.24 & 62.09 & 57.64 & \multirow{2}{*}{+1.92\tiny{$\pm$2.34}} \\
& \cmark & \textbf{73.25} & \textbf{76.12} & \textbf{69.67} & \textbf{62.52} & \textbf{69.95} & \textbf{56.02} & \textbf{63.40} & \textbf{64.95} & \textbf{65.35} & \\
\midrule

\multirow{2}{*}{CRIS~\cite{wang2022cris}} & \xmark & 66.68 & 70.62 & 59.93 & 56.94 & 64.20 & 46.97 & 55.91 & 58.50 & 54.55 & \multirow{2}{*}{+1.73\tiny{$\pm$0.82}} \\
& \cmark & \textbf{68.66} & \textbf{72.82} & \textbf{63.06} & \textbf{57.94} & \textbf{65.25} & \textbf{48.41} & \textbf{58.47} & \textbf{59.07} & \textbf{56.23} & \\
\midrule
\multirow{2}{*}{ReLA~\cite{liu2023gres}} & \xmark &  73.67  & 76.18 & 70.39 & 63.82 & 68.70 & 55.78 & 65.22 & 65.29 & 63.10 & \multirow{2}{*}{+0.97\tiny{$\pm$0.82}} \\
& \cmark & \textbf{74.24} & \textbf{77.11} & \textbf{70.39} & \textbf{65.35} & \textbf{70.55} & \textbf{56.68} & \textbf{65.32} & \textbf{65.73} & \textbf{65.54} & \\
\midrule
\multirow{2}{*}{CGFormer~\cite{tang2023cgformer}} & \xmark &  72.53  & 75.12 & 70.09 & 63.55 & 68.58 & 56.05 & 62.92 & 64.63 & 64.77 & \multirow{2}{*}{+1.04\tiny{$\pm$0.67}} \\
& \cmark & \textbf{73.52} & \textbf{76.07} & \textbf{70.92} & \textbf{64.30} & \textbf{69.58} & \textbf{57.85} & \textbf{65.31} & \textbf{65.07} & \textbf{65.00} & \\
\midrule
\multirow{2}{*}{VPD~\cite{zhao2023unleashing}} & \xmark &  73.46  & 75.31 & 70.23 & 61.41 & 67.98 & 54.99 & 63.12 & 63.59 & 62.38 & \multirow{2}{*}{+1.47\tiny{$\pm$0.85}} \\
& \cmark & \textbf{74.48} & \textbf{76.32} & \textbf{71.51} & \textbf{62.86} & \textbf{69.92} & \textbf{55.56} & \textbf{64.40} & \textbf{64.80} & \textbf{65.89} & \\
\midrule
Average Gain & & \multicolumn{3}{c|}{+1.11\tiny{$\pm$0.79}} & \multicolumn{3}{c|}{+1.21\tiny{$\pm$0.48}} & \multicolumn{2}{c|}{+1.55\tiny{$\pm$0.99}} & {+3.11\tiny{$\pm$2.83}} &  \\
\bottomrule
\end{tabular}}
\end{table*}

\label{sec:exp:effect}


\noindent
\textbf{Datasets.}
We evaluate on four widely-used RIS benchmarks: RefCOCO~\cite{kazemzadeh2014referitgame}, RefCOCO+~\cite{kazemzadeh2014referitgame}, G-Ref~\cite{mao2016generation}, and GRES~\cite{liu2023gres}.
Dataset statistics are summarized in \cref{tab:ris_dataset_stats}.
For RefCOCO and RefCOCO+, we use the train, validation, test A and B partitions. We use both UMD~\cite{yu2016modeling} and Google~\cite{mao2016generation} split for G-Ref dataset.
Refer to \cref{sec:googles_split} for the Google split. 

\vspace{0.1cm} 
\noindent
\textbf{Metrics.}
We adopt three metrics.
First, overall intersection over union is the proportion of the intersection area to the union area across all test samples.
Due to its tendency to favor larger objects, we also use mean intersection over union, representing the average intersection between the prediction and the ground truth for all samples.
Lastly, we report Precsion@$p$, the ratio of samples whose IoU with the ground truth exceeds the threshold $p$, with $p \in$ \{0.5, 0.7, 0.9\}.

\vspace{0.1cm} 
\noindent
\textbf{RIS Models.}
To verify the applicability of our data augmentation method, we experiment with five state-of-the-art RIS models: LAVT~\cite{yang2022lavt}, CRIS~\cite{wang2022cris}, ReLA~\cite{liu2023gres}, CGFormer~\cite{tang2023cgformer} and VPD~\cite{zhao2023unleashing}.
For each method, we compare the overall RIS performance with and without applying our NeMo.
Refer to Appendix \ref{sec:impl} for implementation details.



\begin{figure}[t]
    \includegraphics[width=\linewidth]{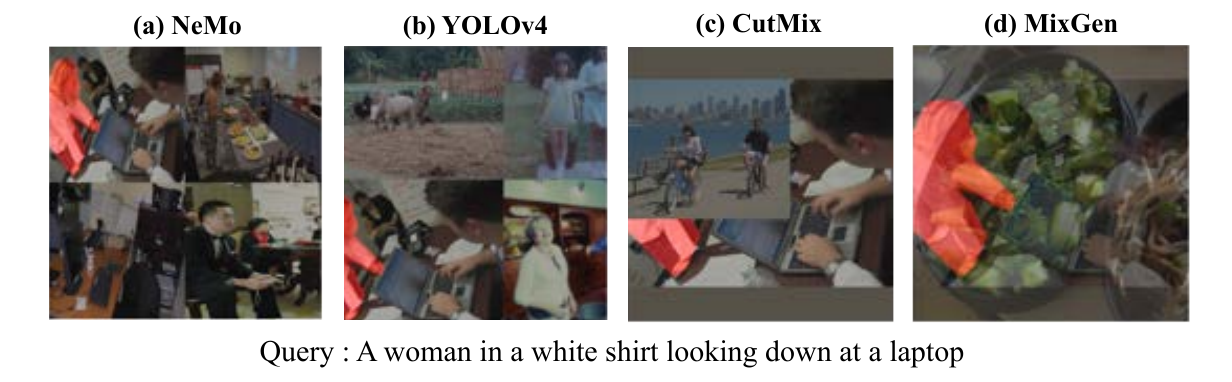}
    \caption{\textbf{Samples from other multi-modal augmentation method.} $2\times 2$ mosaic samples from (a)NeMo, (b)YOLOv4, (c)CutMix, (d)MixGen.}
    \label{fig:other_aug_sample}
\end{figure}

\begin{figure}[t]
  \centering
  \begin{minipage}{.53\textwidth}
    \centering
    \captionof{table}{Comparison to other multi-modal augmentations on G-Ref}
    \label{tab:other_aug}
    \renewcommand{\tabcolsep}{4pt}
    \resizebox{\linewidth}{!}{
    \begin{tabular}{l | c c | c c c }
      \toprule
      {Augmentation}  & \multicolumn{2}{c|}{oIoU} & \multicolumn{2}{c}{Prec (Val)} \\
      Method & Val & Test & 0.5 & 0.7 \\
      \midrule
      {CRIS} & 55.91 & 58.50 & 67.95 & 54.84 \\
      {+YOLOv4~\cite{bochkovskiy2020yolov4}} & 56.22 & 58.55 & 66.94 & 53.54 \\
      {+CutMix~\cite{yun2019cutmix}}& 56.50 & 58.34 & 66.63 & 53.11 \\
      {+MixGen~\cite{hao2023mixgen}} & 53.62 & 55.85 & 64.37 & 51.28 \\
      {+NeMo (Ours)} & \textbf{58.47} & \textbf{59.07} & \textbf{70.01} & \textbf{56.60} \\
      \bottomrule
    \end{tabular}}
  \end{minipage}%
  \hfill
  \begin{minipage}
  {.43\textwidth}
  \centering
  \caption{\textbf{mIoUs for the different number of negative objects.} }
  \label{fig:n_obj_miou}
  \includegraphics[width=\linewidth]{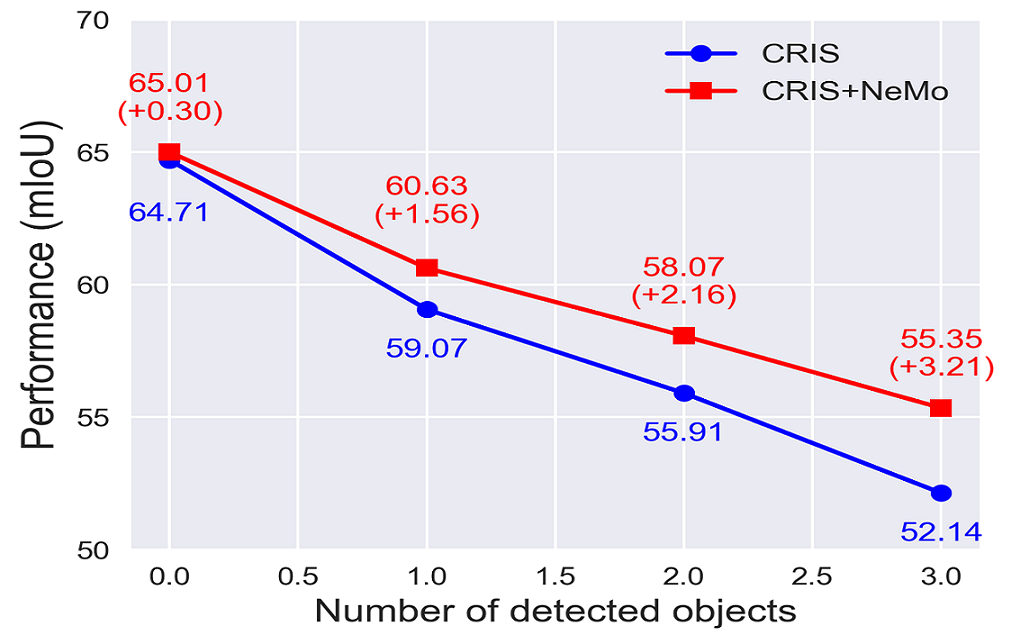}
    \end{minipage}
\end{figure}

\subsection{Effectiveness of the Proposed Method}

\cref{tab:performance_metrics} compares the performance of various RIS models with and without applying our method
in Overall IoU (see \cref{sec:miou_result} for Mean IoU).
We observe that NeMo improves the performance under all settings, across all datasets regardless of the RIS model.
This indeed reveals that a bottleneck has also been in data, not just the modeling aspect, and our method effectively provides ambient training examples with a curated complexity.

An interesting observation is that the degree of improvement differs by the datasets.
On average, we observe a larger performance boost on a more complex dataset (G-Ref) than simpler ones (RefCOCO and RefCOCO+).
Compared to the other two, G-Ref contains nearly twice many objects and three times longer queries (see \cref{tab:ris_dataset_stats}).
G-Ref benefits more from NeMo because of its intricate referring expressions and visually dense scenes.
Similarly, on GRES, extension of RefCOCO+ with no-target and multi-target queries, NeMo shows significant performance gain.
The complex expressions in GRES amplifies the challenge of correctly identifying target(s), and the result validates the robustness of our method in handling more complex referring scenarios.
In contrast, our approach provides less impact on the relatively simpler datasets, because they do not require a nuanced understanding to differentiate similar objects.
Nevertheless, we still notice some improvements even on these simpler datasets, indicating that NeMo is still beneficial and does not affect negatively.

\subsection{Comparison with Other Augmentation Methods}

\cref{tab:other_aug} compares different multi-modal augmentation methods on G-Ref. 
We observe that most methods degrade performance, showing that other augmentation methods are not suitable for the RIS task.
\cref{fig:other_aug_sample} illustrates how they can fail for RIS; YOLOv4 and CutMix often lose or obstruct referents after cropping or overlaying the image.
MixGen also underperforms, likely due to difficulties understanding the whole scene of the original image while interpolated.
These results endorse NeMo as a suitable augmentation approach for the RIS task.


\begin{figure}[t]
  \centering
  \begin{minipage}{.53\textwidth}
    \centering
    \captionof{table}{Performance over various sentence lengths on G-Ref UMD test split.}
    \label{tab:sent_len_exp}
    \renewcommand{\tabcolsep}{4pt}
    \resizebox{\linewidth}{!}{
    \begin{tabular}{lc|cccc}
        \toprule
        \multirow{2}{*}{RIS model} & \multirow{2}{*}{NeMo} & \multicolumn{4}{c}{Length of $T$} \\
        & & 1-5 & 6-7 & 8-10 & 11-20 \\
        \midrule
        \multirow{2}{*}{LAVT\cite{yang2022lavt}} & \xmark & 63.95 & 63.46 & 63.03 & 63.00 \\
         & \cmark & \textbf{66.50} & \textbf{65.39} & \textbf{64.40} & \textbf{64.72} \\
        \midrule
        \multirow{2}{*}{CRIS\cite{wang2022cris}} & \xmark & 58.91 & 56.41 & 55.29 & 57.33 \\
         & \cmark & \textbf{60.77} & \textbf{57.17} & \textbf{57.05} & \textbf{58.35} \\
        \midrule
        \multirow{2}{*}{ReLA\cite{liu2023gres}} & \xmark & \textbf{66.67} & 64.95 & \textbf{63.82} & 65.95 \\
        & \cmark & 66.63 & \textbf{65.00} & 63.75 & \textbf{67.26} \\
        \midrule
        \multirow{2}{*}{CGFormer\cite{tang2023cgformer}} & \xmark & 65.85 & 65.12 & \textbf{64.33} & 63.87 \\
         & \cmark & \textbf{66.30} & \textbf{65.44} & 63.98 & \textbf{64.98} \\
        \midrule
        \multirow{2}{*}{VPD\cite{zhao2023unleashing}} & \xmark & \textbf{67.53} & 66.12 & 65.49 & 67.44 \\
         & \cmark & 66.30 & \textbf{66.86} & \textbf{67.33} & \textbf{68.12} \\
        \bottomrule
    \end{tabular}}
  \end{minipage}%
  \hfill
  \begin{minipage}{.43\textwidth}
  \centering
  \caption{\textbf{mIoUs for various object sizes.} Sizes are binned such that each bin contains 10\% of the test set.}
  \label{fig:obj_scale_scores}
  \includegraphics[width=\linewidth]{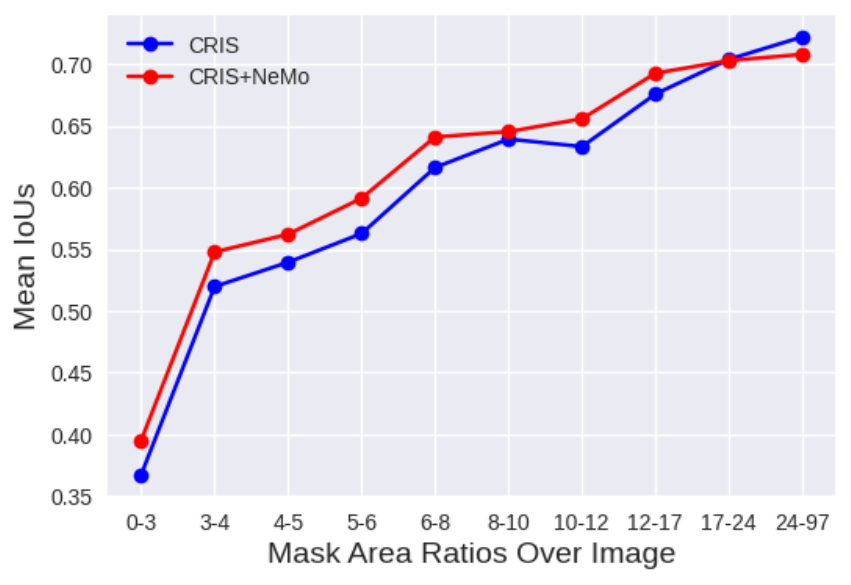} 
    \end{minipage}
\end{figure}

\subsection{Detailed Analysis of the Proposed Method}
\label{sec:exp:analysis}



\noindent
\textbf{Performance on Challenging Scenarios.}
RIS task is challenging when it cannot be easily resolved through class or positional keywords alone. This occurs when there exist multiple objects of the same class as the referent, where we expect NeMo to be particularly effective.
To evaluate this systematically, we compare the performance with varying number of negative objects within the image.
We first detect objects in the scene with a pre-trained detector, and identify same-class objects that closely overlap with the target to count negative objects.
\cref{fig:n_obj_miou} shows that the performance gap between with and without NeMo gets larger with more negative objects in the image.
This indicates that NeMo performs better on challenging samples as our expectation.

\begin{figure}[t]
  \centering
  \includegraphics[width=\linewidth]{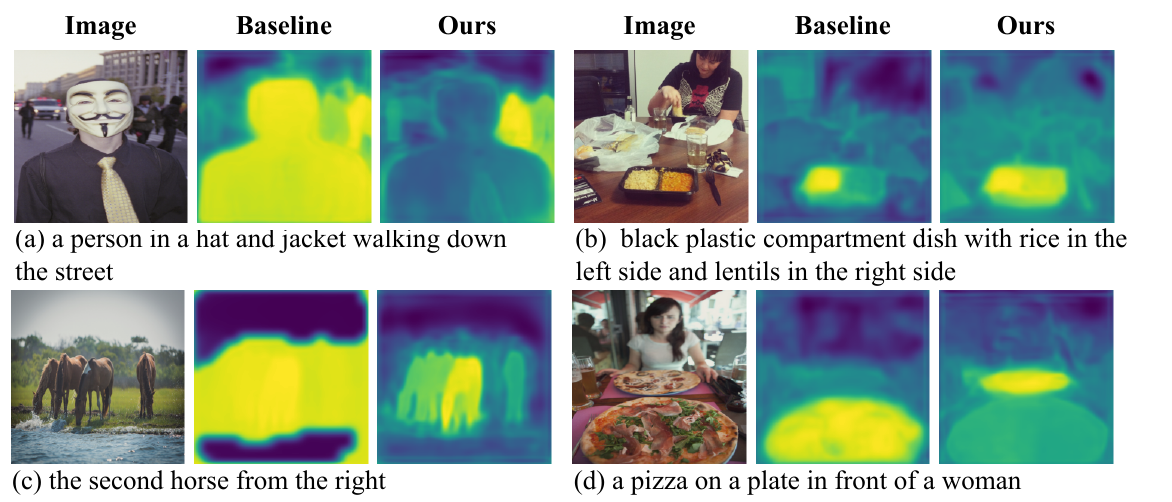} 
  \caption{Visualization of activation maps with and without NeMo on CRIS}
  \label{fig:activation_map}
\end{figure}

\vspace{0.1cm} 
\noindent
\textbf{Robustness on Object Scale.}
We evaluate the impact of our method across various object sizes in \cref{fig:obj_scale_scores}.
Improvement is observed in most cases, especially for smaller objects.
This can be attributed to the wider range of object scales seen during training by integrating objects both in the original and 1/4 size.
\cref{fig:activation_map}(a) demonstrates that our method successfully detects a blurred and small ``person'' in the background, positioned behind the most prominent person in the image.
Overall, we observe clearer boundaries in the final activation maps with NeMo, commonly illustrated in \cref{fig:activation_map}.

\vspace{0.1cm}
\noindent
\textbf{Complexity of Referring Expression.}
Following \cite{liu2017recurrent, kim2022restr}, we measure how our method behaves depending on sentence lengths.
\cref{tab:sent_len_exp} shows that NeMo is effective across all sentence lengths, even with longer complex ones.
NeMo also helps capture important linguistic cues for grounding.
In \cref{fig:activation_map}(b), our method segments the entire dish while the baseline only detects the left half without fully understanding the query describing the right half.

\begin{figure}[t]
  \centering
  \small
  \begin{minipage}{0.53\linewidth}
    \centering
    \captionof{table}{Effect of FP on RefCOCO}
    \label{tab:fp_exp}
    \footnotesize
    \renewcommand{\tabcolsep}{5pt}
    \resizebox{\linewidth}{!}{
    \begin{tabular}{l | c c c | c c c}
      \toprule
      {}  & \multicolumn{3}{c|}{oIoU} & \multicolumn{2}{c}{Prec (Val)} \\
      Method  & Val & TestA & TestB & 0.5 & 0.7 \\
      \midrule
      {CRIS}  & 66.68 & 70.62 & 59.93 & 80.89 & 69.23\\
      {Ours} & \textbf{68.66} & \textbf{72.82} & \textbf{63.33} & \textbf{82.87} & \textbf{70.63} \\ 
      {Ours wo/ FP}  & 68.30 & 72.41 & 63.26 & 81.96 & 70.18\\
      \bottomrule
    \end{tabular}}
  \end{minipage}%
  \hfill
  \begin{minipage}{0.45\linewidth}
  \centering
  \caption{\textbf{} Performance gain for queries with and without positional keywords.}
  \label{fig:pos_sent_len}
  \includegraphics[width=\linewidth]{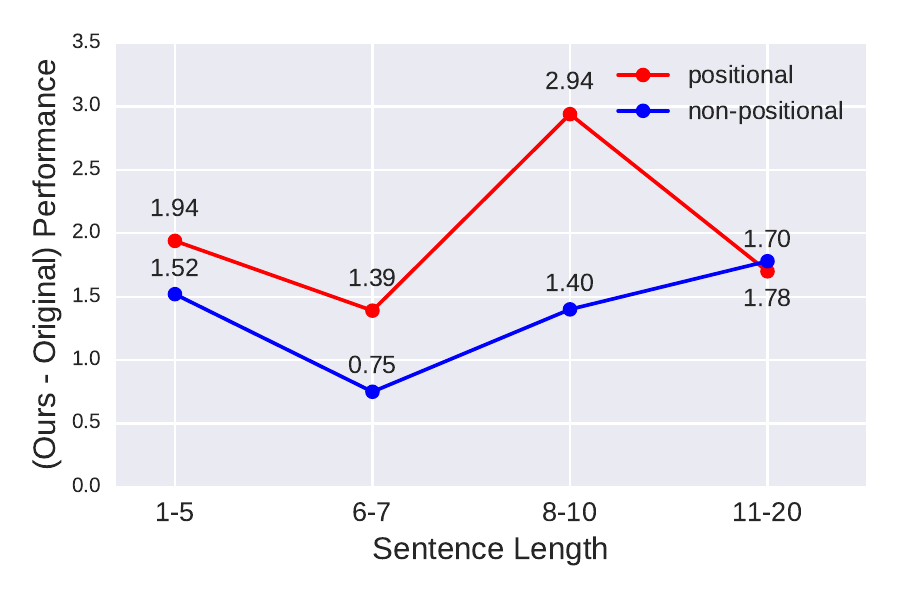} 
  \end{minipage}
\end{figure}

\vspace{0.1cm} 
\noindent
\textbf{Positional Understanding.}
\cref{fig:pos_sent_len} demonstrates that NeMo exhibits stronger improvements especially for queries with positional keywords even when the queries are long and complex. Along with the improved visual and textual understanding with NeMo, the model is exposed to various samples in mosaic not confined to the specific local positions, which requires deeper positional understanding.
As seen in \cref{fig:activation_map}(b,d), our method captures objects more accurately in scenarios involving directional expressions, indicating improved understanding both in absolute and relative positions.
In \cref{fig:activation_map}(c), it is apparent that our method yields an output with more distinct shape for ``the second'' horse.


\vspace{0.1cm}
\noindent
\textbf{Effect of False Positives.}
Even after tuning $\tau$, one might concern non-zero probability of FPs.
To address this, we conduct a rule-based experiment to restrict the location of the positive image, if the text contains positional keywords.
For example, images labeled with `left' are forced to be in one of the left quadrants.
This experiment is carried out on RefCOCO, featured by simpler texts.
As shown in \cref{tab:fp_exp}, performance gap between with and without the constraint turns out to be minimal, indicating that $\tau$ and $K$ is effective in mitigating FPs.
Intriguingly, NeMo without the constraint performs the best.
We interpret this as 1) NeMo hardly creates FPs with well-tuned $\tau$ and $K$, and 2) even if some FPs exist, the benefit from allowing more diversity of augmented images is bigger.
See more discussion in \cref{sec:fp_exp}.

\subsection{Ablation Study}
\label{sec:exp:ablation}


\begin{figure}[t]
  \centering
  \begin{minipage}{.48\textwidth}
    \centering
    \captionof{table}{Ablation on $\gamma$}
    \renewcommand{\tabcolsep}{5pt}
    \resizebox{0.98\linewidth}{!}{
    \label{tab:ablation_gamma}
    \footnotesize
    \begin{tabular}{l|ccccc}
      \toprule
      $\gamma$  & P@0.5 & P@0.7 & P@0.9 & oIoU & mIoU \\
      \midrule
      0.8 & 69.61 & 56.62 & 14.64 & 57.31 & 60.62 \\
      0.6 & 69.00 & 57.09 & 16.06 & \textbf{58.15} & \textbf{60.90} \\
      0.5 & 69.51 & \textbf{57.23} & \textbf{16.38} & 57.81 & 60.85 \\
      0.4 & 69.38 & 56.33 & 15.85 & 57.40 & 60.88 \\
      0.2 & \textbf{70.16} & 56.46 & 15.38 & 57.38 & 60.62 \\
      \bottomrule
    \end{tabular}}
  \end{minipage}%
  \hfill
  \begin{minipage}{.48\textwidth}
    \centering
    \caption{Ablation on $K$}
    \label{fig:ablation_K}
    \includegraphics[width=\linewidth]{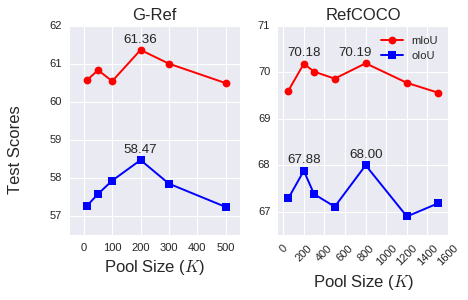}
  \end{minipage}
\end{figure}

\begin{table}[t]
\centering
\caption{Ablation on $\tau$}
\label{tab:ablation_tau}
\footnotesize
\renewcommand{\tabcolsep}{6pt}
\resizebox{0.82\linewidth}{!}{
  \begin{tabular}{l|ccccc|ccccc}
    \toprule 
    \multirow{2}{*}{$\tau$} & \multicolumn{5}{c|}{G-Ref(Val) $K=200$} & \multicolumn{5}{c}{RefCOCO(Val) $K=800$} \\
    & P@0.5 & P@0.7 & P@0.9 & oIoU & mIoU & P@0.5 & P@0.7 & P@0.9 & oIoU & mIoU \\
    \midrule
    1.00 & 69.14 & 55.76 & 15.09 & 58.06 & 60.73 & 81.90 & 69.36 & 19.96 & \textbf{68.04} & 70.34 \\
    0.85 & 69.47 & 56.29 & 15.08 & 57.57 & 60.50 & \textbf{82.14} & \textbf{71.04} & \textbf{20.07} & 68.01 & \textbf{70.71} \\
    0.75 & \textbf{70.01} & \textbf{56.60} & \textbf{15.89} & \textbf{58.47} & \textbf{61.36} & 81.35 & 70.23 & 19.81 & 68.00 & 70.19\\
    0.60 & 69.63 & 56.17 & 14.07 & 57.92 & 60.36 & 81.66 & 70.02 & 19.31 & 67.74 & 70.29 \\
    \bottomrule
  \end{tabular}}
\end{table}

We conduct an ablation study to break down the contributions of individual components of NeMo.
All experiments are conducted with CRIS~\cite{wang2022cris} on the UMD validation set of G-Ref and UNC validation set of RefCOCO.

\vspace{0.1cm} 
\noindent
\textbf{Augmentation Ratio $\gamma$.}
Recall from \cref{sec:neg_img_selection} that we apply the proposed augmentation with the probability of $\gamma$ to each training sample. When NeMo is not applied, original single images are used for training as the model eventually infers on the single images. Before adjusting the difficulty of mosaics, we conduct this experiment at the median level of difficulty by setting $K = |\mathcal{D}|/2$. 
~\cref{tab:ablation_gamma} indicates that best performance is obtained when $\gamma$ is between 0.5 to 0.6, while performance slightly drops with $\gamma$ too large or too small.

\vspace{0.1cm} 
\noindent
\textbf{Negative Image Pool Size $K$.}
We explore the optimal pool size of the candidate negative images, $K$, on both G-Ref and RefCOCO.
Adjusting $K$ tunes not just the number, but also the minimum relevance of the negative images to be considered.
The other hyperparameters are set to the default values, $\tau = 0.75$ and $\gamma = 0.6$.
\cref{fig:ablation_K} reveals that both exceedingly low and high $K$ are suboptimal.
With too small $K$, only mostly similar images may remain, creating FP and FN frequently.
In contrast, an extremely large $K$ brings little additional challenge, getting closer to the uniform selection.
This confirms that a moderate level of challenge in the negative samples is most beneficial.
Besides, the optimal $K$ values differ significantly by the datasets: 200 for G-Ref and 800 for RefCOCO.
In a simpler dataset, allowing less visually similar images would enhance overall performance by minimizing the chances to cause FP/FN.


\vspace{0.1cm} 
\noindent
\textbf{Upper-bound Threshold $\tau$.}
We explore the optimal $\tau \in \{0.6, 0.75, 0.85, 1.0\}$ where $K$ is fixed to the best value found for each dataset above.
\cref{tab:ablation_tau} shows that a moderate $\tau$ around 0.75 to 0.85 yields the best performance.
With a larger $\tau$, less images are filtered out, causing more chances for FP/FN. With too small $\tau$, it becomes closer to random sampling again.
We further explore the cross-effect of $\tau$ and $K$ and mosaic design choices in~\cref{sec:further_abl}.

\begin{figure}[t]
    \centering
    \includegraphics[width=0.97\linewidth]{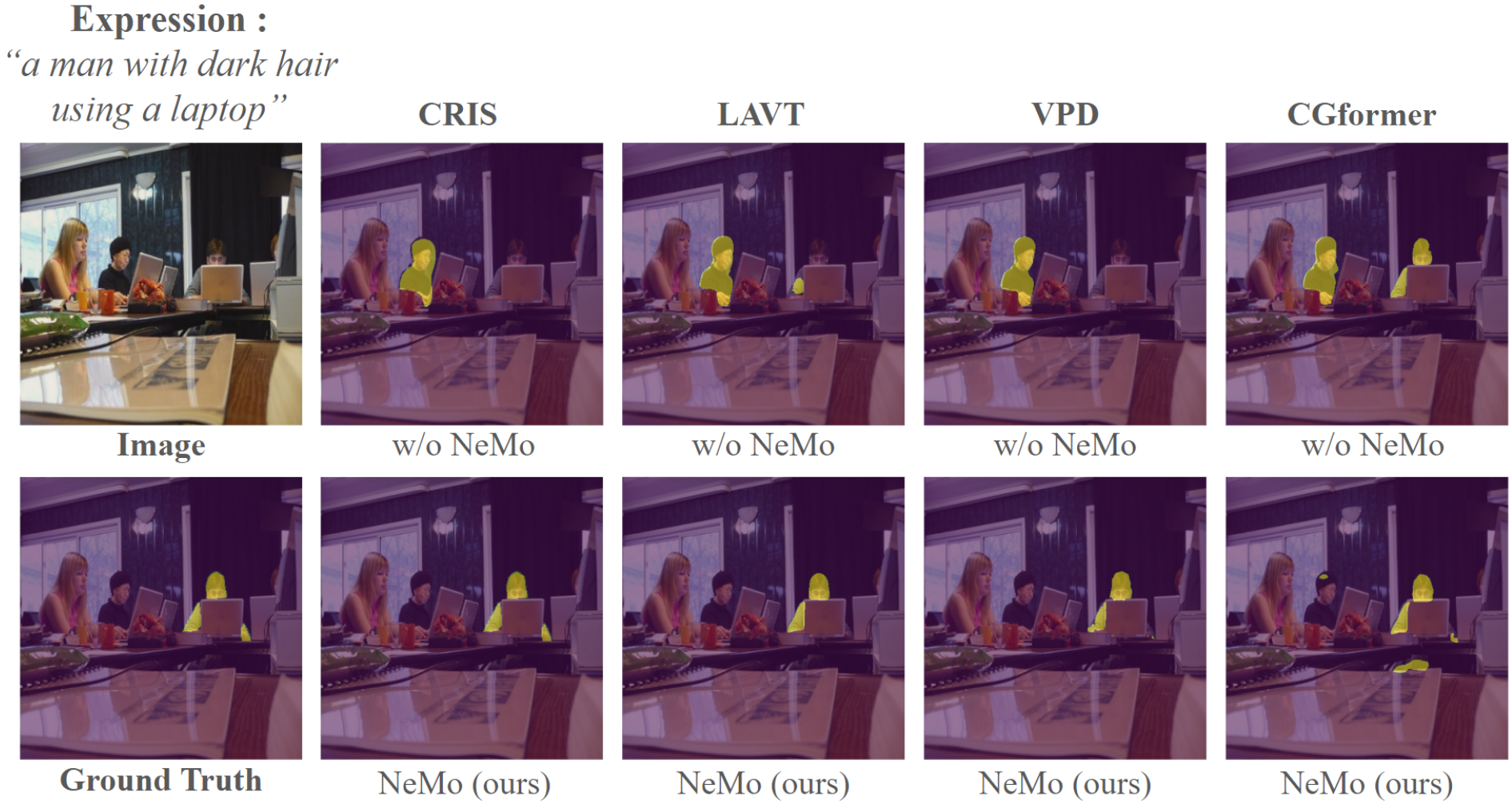}
    \caption{Visualization of predictions from the G-Ref test set. }
    \label{fig:qual_res}
\end{figure}

\section{Conclusion} 
\label{sec:conclusion}

This paper proposes NeMo, an advanced mosaic augmentation method that exposes an RIS model to more intricate and challenging examples, thereby enhancing its visual and linguistic understanding for locating and segmenting the referent.
NeMo brings consistent performance improvement over various state-of-the-art RIS models on multiple datasets, especially on datasets with higher visual-linguistic complexity.
Although NeMo shows consistent improvement, data augmentation in this context still remains a relatively unexplored territory.
More sophisticated methodologies, such as an object-level parsing, could potentially further enhance retrieval, but we leave this for future research.

\vspace{0.1cm} \noindent
\textbf{Limitations.}
Our method performs well when the dataset contains relatively homogeneous images. If it contains images from diverse domains (\textit{e.g.}, X-ray, sketches, or satellite images), mosaic creation may result in unnatural combinations, potentially degrading the performance.



\clearpage
\section*{Acknowledgements}

We appreciate Jeongwoo Shin for insightful discussion.
This work was supported by the
New Faculty Startup Fund from Seoul National University, by Samsung Electronics Co., Ltd (IO230414-05943-01, RAJ0123ZZ-80SD), by Youlchon Foundation (Nongshim Corp.), and by National Research Foundation (NRF) grants (No.
2021H1D3A2A03038607/50\%, RS-2024-00336576/10\%, RS-
2023-00222663/5\%) and Institute for Information \& communication Technology Planning \& evaluation (IITP) grants (No. RS-2024-00353131/25\%, RS-2022-II220264\\
/10\%), funded by the government of Korea.


%
%
\bibliographystyle{splncs04}
\bibliography{main}

\newpage
\clearpage

\appendix
\pagenumbering{roman}
\renewcommand\thetable{\Roman{table}}
\renewcommand\thefigure{\Roman{figure}}
\setcounter{table}{0}
\setcounter{figure}{0}

\setcounter{page}{1}

\section*{Appendix}


\section{Easy \& Hard Examples}
\label{sec:easy_hard_examples}
In our analysis in \cref{sec:intro}, we categorize examples from the G-Ref UMD test set into `easy' and `hard' samples to illustrate the varying levels of challenge present within the same dataset. 

\begin{figure*}
\centering
\includegraphics[width=\linewidth]{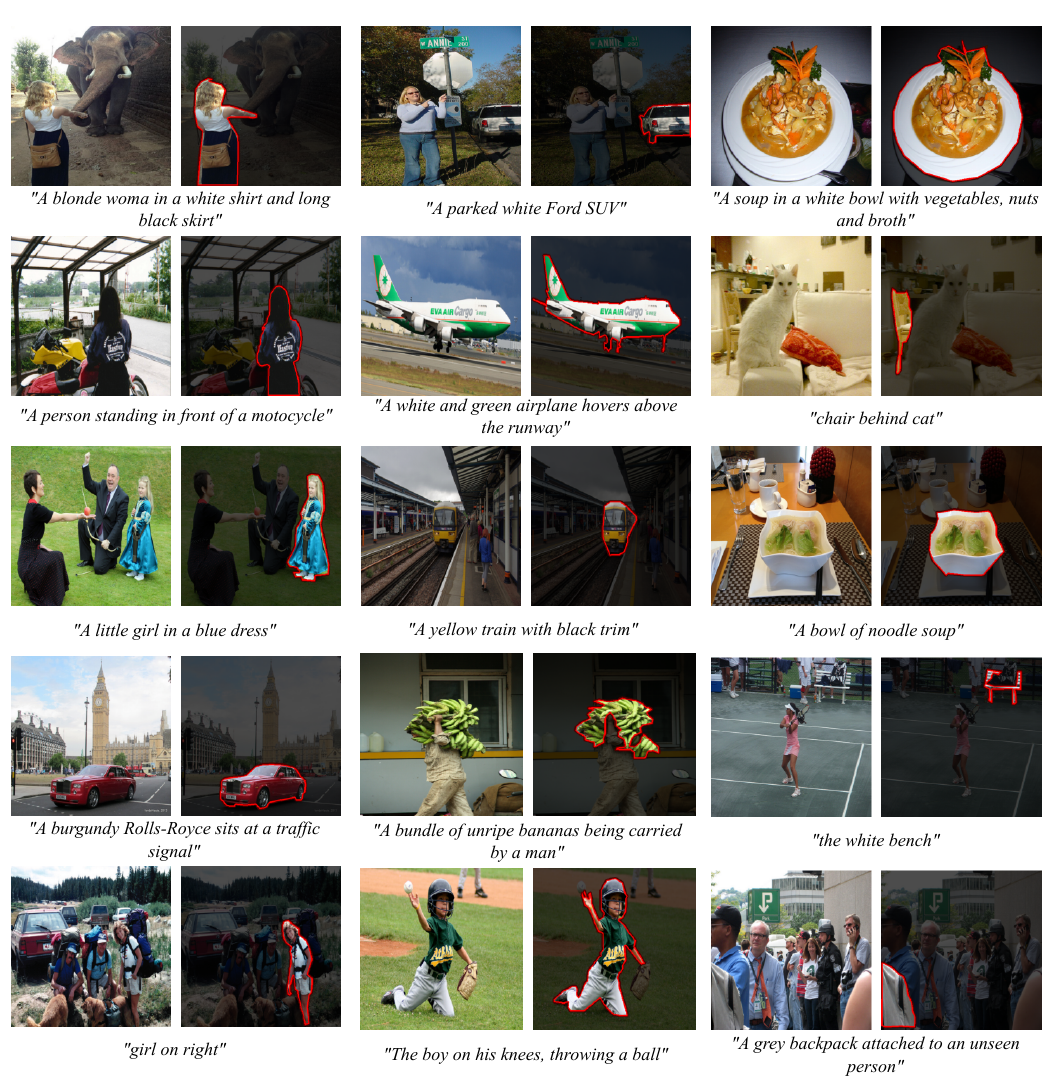} 
\caption{Easy samples from G-Ref test split}
\label{fig:easy_samples} 
\end{figure*}

The `Easy' samples in \cref{fig:easy_samples} feature scenes with a single object in a category, making them straightforward to identify.
For example, the `airplane' in the second image of the second row is the only object in its category, clearly aligning with the referring text.
In other scenes like the `SUV' (row 1, column 2) and the `bench' (row 4, column 3), despite multiple objects in view, the target object stands out as the only one in their respective categories.
These samples highlight instances within the dataset where the target can be readily determined by the subject alone.

\begin{figure*}
\centering
\includegraphics[width=\linewidth]{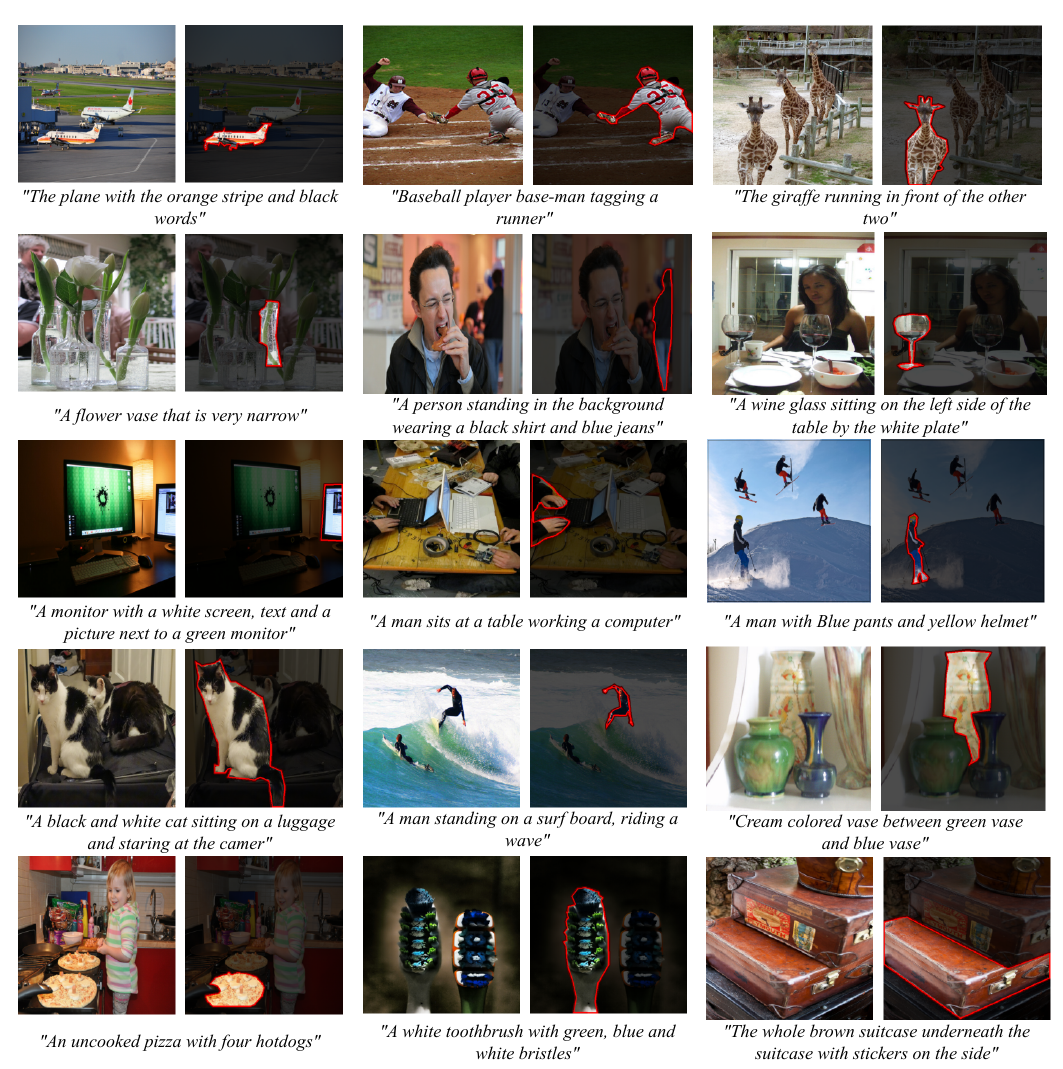} 
\caption{Hard samples from G-Ref test split}
\label{fig:hard_samples} 
\end{figure*}

On the other hand, `hard' samples feature multiple objects from the same category, requiring a nuanced understanding of context to pinpoint the target.
For instance, in \cref{fig:hard_samples}, to find the right `giraffe' in the first row's third image, one must discern its position relative to the `other two' giraffes.
In addition to positional cues, some instances demand comprehension of descriptive phrases or modifiers to correctly identify objects.
In the second row's first image, finding the `flower vase' depends on grasping the word `narrow', and locating `a man' in the third row's third image necessitates understanding the action `working on a computer'.

These examples in the \cref{fig:easy_samples} and \cref{fig:hard_samples} demonstrate how the complexity of a scene affects the difficulty of object identification.

\section{Implementation Details}
\label{sec:impl}
We cross-validate $\gamma$ within $[0.2, 0.8]$ and choose 0.6 as the default.
We empirically set ($\tau$, $K$) to (0.75, 200) for G-Ref and (0.85, 800) for RefCOCO(+) (see \cref{sec:exp:ablation}).
For the RIS models, we employ Swin Transformer Base~\cite{liu2021swin} and BERT~\cite{devlin2018bert} for the visual and text backbone, respectively.
In the case of CRIS, we use ResNet50~\cite{he2016deep} and Transformer~\cite{vaswani2017attention} backbones.
We use the AdamW~\cite{yang2021bottom} optimizer with learning rates of $10^{-5}$ for ReLA, $5 \cdot10^{-5}$ for VPD and LAVT, and $10^{-4}$ for CGFormer and CRIS.
The training epochs remain consistent with the original papers.
Batch size is set to 32 for LAVT and VPD, and 64 for CGFormer and CRIS. 
Unless noted above, we follow the settings from the original papers.

\newpage
\section{Performance Comparison in Mean IoU}
\label{sec:miou_result}

\cref{tab:miou_scores} compares the performance of RIS models with and without our NeMo applied, in terms of Mean IoU (mIoU).
As similarly in the case of oIoU in \cref{tab:performance_metrics}, applying NeMo consistently improves the RIS performance across most models and datasets.

\begin{table}[h!]
\centering
\caption{mIoU scores for recent RIS models.}
\label{tab:miou_scores}
\setlength{\tabcolsep}{4pt}
\resizebox{1\linewidth}{!}{
\begin{tabular}{lc|ccc|ccc|cc|cc}
\toprule
\multirow{2}{*}{Method} & \multirow{2}{*}{NeMo} & \multicolumn{3}{c|}{RefCOCO(UNC)} & \multicolumn{3}{c|}{RefCOCO+(UNC+)} & \multicolumn{2}{c|}{G-Ref (UMD)} & \multicolumn{1}{c}{GRES} \\
& & Val & TestA & TestB & Val & TestA & TestB & Val(U) & Test(U) & Val\\
\midrule
\multirow{2}{*}{LAVT} & \xmark & 74.31 & 76.63 & 70.61 & 64.99 & 70.98 & 59.00 & 63.34 & 63.62 & 58.40 \\
& \cmark & \textbf{75.03} & \textbf{77.24} & \textbf{71.87} & \textbf{66.00} & \textbf{71.75} & \textbf{60.35} & \textbf{66.19} & \textbf{66.67} & \textbf{66.94} \\
\midrule
\multirow{2}{*}{CRIS} & \xmark & 69.52 & 72.72 & 64.70 & 61.39 & 67.10 & 52.48 & 59.35 & 59.39 & 49.37\\
& \cmark & \textbf{70.93} & \textbf{74.19} & \textbf{66.16} & \textbf{62.40} & \textbf{68.52} & \textbf{53.44} & \textbf{61.36} & \textbf{61.38} & \textbf{51.69} \\
\midrule
\multirow{2}{*}{ReLA} & \xmark & 75.61 & 77.79 & 72.82 & 66.67 & 70.71 & 60.35 & 68.02 & \textbf{68.29} & 63.60 \\
& \cmark & \textbf{75.76} & \textbf{78.04} & \textbf{73.18} & \textbf{68.94} & \textbf{73.11} & \textbf{61.63} & \textbf{68.38} & 68.04 & \textbf{70.20} \\
\midrule
\multirow{2}{*}{CGFormer} & \xmark & 75.28 & 77.32 & 72.51 & 67.81 & 71.90 & 61.03 & 66.62 & 67.53 & 65.26 \\
& \cmark & \textbf{75.82} & \textbf{77.57} & \textbf{73.47} & \textbf{68.62} & \textbf{72.75} & \textbf{62.42} & \textbf{68.19} & \textbf{67.80} & \textbf{65.94} \\
\midrule
\multirow{2}{*}{VPD} & \xmark & 75.67 & 77.39 & 73.23 & 67.98 & 71.82 & 60.39 & 66.42 & 66.75 & 65.37 \\
& \cmark & \textbf{76.42} & \textbf{78.54} & \textbf{74.29} & \textbf{68.49} & \textbf{73.24} & \textbf{60.77} & \textbf{67.41} & \textbf{67.79} & \textbf{70.13} \\
\bottomrule
\end{tabular}}
\end{table}

\section{Performance on G-Ref Google Split}
\label{sec:googles_split}
\cref{tab:experiments} presents additional performance comparison on the G-Ref Google split validation set, providing additional context to the IoU performance metrics outlined in \cref{tab:performance_metrics} of the main manuscript.
Again, we consistently observe improved performance with our NeMo across various models.
Overall IoU metric shows notable improvement with respective gains of 2.55\% for LAVT, 3.43\% for CRIS, 1.61\% for VPD, and 2.06\% for CGFormer.
These results underscore the robustness of our augmentation method across different architectures.

\begin{table}[h]
  \centering
  \caption{Overall IoU on G-Ref Google split validation set.}
  \label{tab:experiments}
  \footnotesize
  \begin{tabular}{c|c|c|c|c|c}
    \toprule
    Mosaic & LAVT~\cite{yang2022lavt} & CRIS~\cite{wang2022cris} & ReLA~\cite{liu2023gres} & VPD~\cite{zhao2023unleashing} & CGFormer~\cite{tang2023cgformer} \\
    \midrule
    \xmark & 59.41  & 52.26 & 60.76 & 60.23 & 61.20 \\
    \cmark & 61.96 & 55.69 & 63.89 & 61.84 & 63.26 \\
    \midrule
    Diff & \textbf{+2.55} & \textbf{+3.43} & \textbf{+3.13} & \textbf{+1.61} & \textbf{+2.06}\\
    \bottomrule    
  \end{tabular}
\end{table}

\section{Finding the Optimal $\tau$ and $K$ on RefCOCO+}
We additionally conduct an experiment to find the optimal values for $\tau$ and $K$ in RefCOCO+, using methodologies similar to those in our previous ablation studies for G-Ref and RefCOCO in \cref{sec:exp:ablation}.

\begin{figure}[h]
  \centering
   \begin{minipage}{.48\textwidth}
    \centering
    \includegraphics[width=\linewidth]{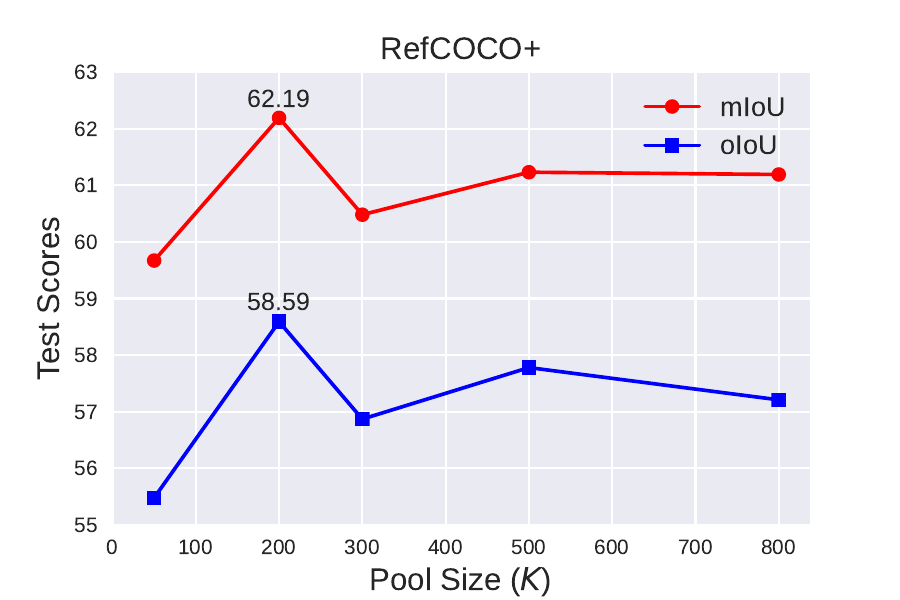}
    \caption{Ablation study on $K$}
    \label{fig:ablation_K_refcoco+}
  \end{minipage}
  \begin{minipage}{.48\textwidth}
    \centering
    \captionof{table}{Ablation study on $\tau$}
    \renewcommand{\tabcolsep}{5pt}
    \resizebox{0.98\linewidth}{!}{
    \label{tab:ablation_tau_refcoco+}
    \footnotesize
    \begin{tabular}{l|ccccc}
      \toprule
      $\tau$  & P@0.5 & P@0.7 & P@0.9 & oIoU & mIoU \\
      \midrule
      1.00 & 70.05 & 58.41 & 14.98 & 56.72 & 60.83  \\
      0.85 &  71.10 & 59.77 & 15.34 & \textbf{58.40} & 61.57 \\
      0.75 &  \textbf{71.41} & \textbf{60.56} & \textbf{15.45} & 57.91 & \textbf{61.80}  \\
      0.60 &  69.21 & 58.30 & 14.77 & 56.92 & 60.60  \\
      \bottomrule
    \end{tabular}}
  \end{minipage}%
  \hfill
 
\end{figure}

\vspace{0.1cm} \noindent
\textbf{Negative Image Pool Size $K$.} For RefCOCO+, we search the ideal size of the negative image pool, keeping other hyperparameters fixed ($\tau = 0.75$ and $\gamma = 0.6$).
Our results in~\cref{fig:ablation_K_refcoco+} reveal that the best performance comes with a pool of size $K=200$.
This is significantly smaller than the $K=800$ for RefCOCO.
This difference arises mainly form the unique characteristics of the two datasets; unlike RefCOCO, which relies heavily on geometric expressions like ``left'' and ``right'', RefCOCO+ focuses more on attribute-based descriptions.
This shift results in fewer false positives and negatives with a smaller $K$ value in RefCOCO+, leading to superior performance.

\vspace{0.1cm} \noindent
\textbf{Upper-bound Threshold $\tau$.}
We focus on identifying the optimal $\tau$ on RefCOCO+, considering values within ${0.6, 0.75, 0.85, 1.0}$, while keeping $K$ fixed at its best value found above.
As indicated in~\cref{tab:ablation_tau_refcoco+}, we find that a moderate $\tau$, specifically in the range of 0.75 to 0.85, leads to the most favorable performance.

\section{NeMo Examples with Various $\tau$ and $K$}
\label{sec:nemo_more_result}
\begin{figure*}
\centering
\includegraphics[width=0.85\linewidth] {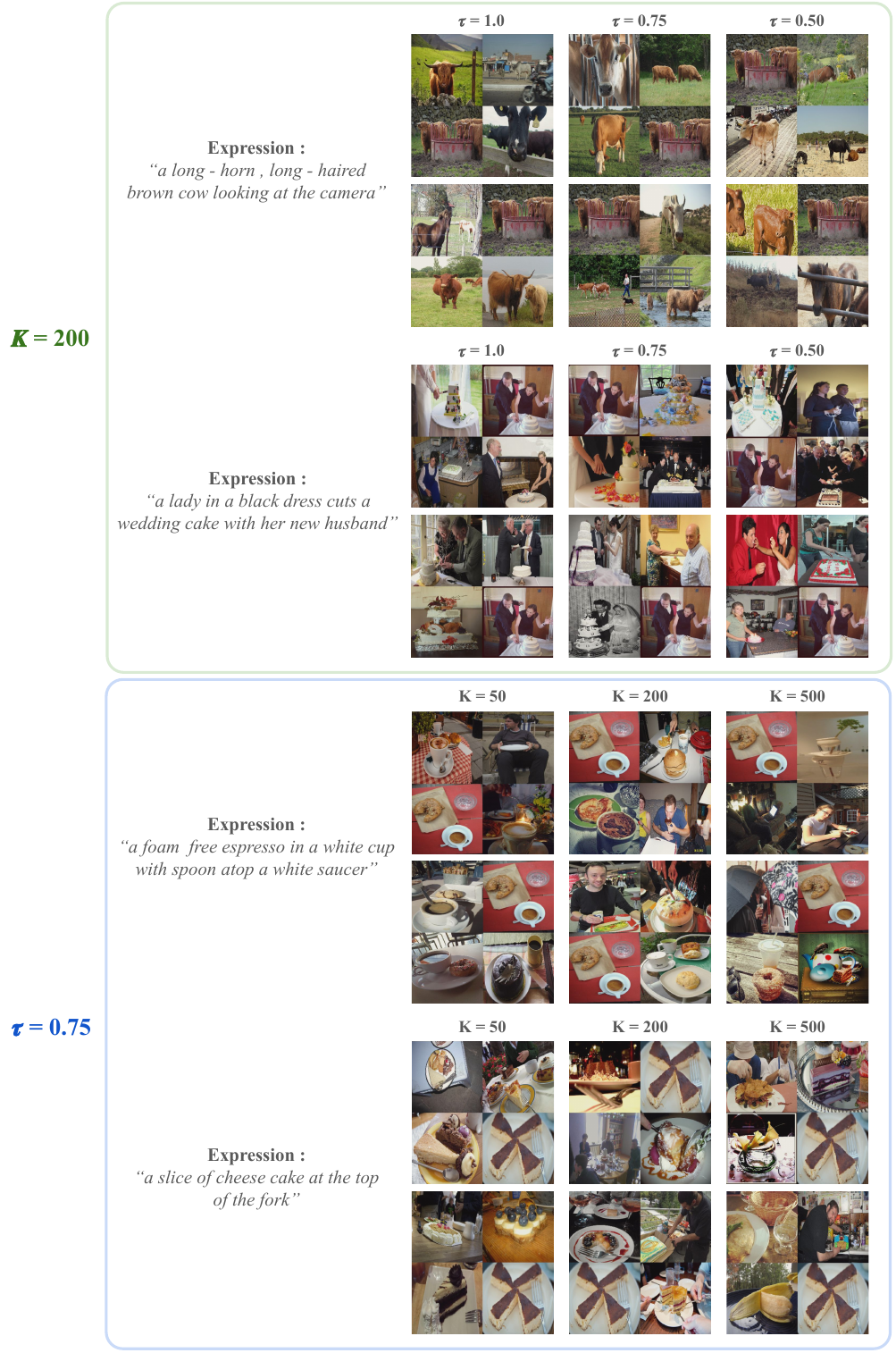}
\caption{NeMo examples with various $\tau$ and $K$.}
\label{fig:tau_k_examples}
\end{figure*}

The key point of NeMo is to filter candidates for negative images of appropriate difficulty based on $\tau$ and $K$, the threshold for the upper bound and the candidate pool size for the lower-bound, respectively.
In \cref{fig:tau_k_examples}, we present augmented results on varying $\tau$ and $K$ to qualitatively demonstrate their effects. 

First, we fix $K$ to be 200 and change $\tau$ to 1.0, 0.75, and 0.50.
When $\tau=1.0$, we observe that most retrieved images are perfectly described by the text, `a long-haired cows looking at the camera' and `ladies in a black dress cutting a cake with a husband', making indistinguishable which one is the intended true image.
With $\tau=0.5$, the retrieved images tend to show more diverse images deviated from the target text.
On the contrary, when $\tau=0.75$, mosaics contain images of similar or same categories to the referents such as cows and a long-haired brown cow with a long horn in the first case.
The second case has images of cakes, dresses and a scene of a lady with a husband.
Nevertheless, the target objects are still distinguishable by the textual cues, ``looking at the camera'' and ``a black dress, cuts a wedding cake'', respectively.   

Next, we change $K$ to 50, 200, and 500 with $\tau$ fixed to 0.75.
When $K$ is small (\emph{e.g.}, 50), the negative pool contains too similar images, resulting in highly ambiguous scenes as seen in examples of ``espresso in a white cup'' and ``cheese cake''.
On the other hand, augmentation with a large $K$ (\emph{e.g.}, 500) creates mosaics whose negative images are irrelevant to the given referring expression.
For instance, an image of a woman with a laptop is retrieved at the expression related to ``espresso''.
With $K=200$, negative images have a moderate difficulty, containing coffee, cakes, and forks.


\section{Further Ablation Study}
\label{sec:further_abl}

\subsection{Combining $\tau$ and \(K\)}
We further delve into the interplay between $\tau$ and \(K\) on G-Ref.
For each, we use either the optimal choice ($\tau = 0.75$ and $K = 200$) in the previous ablation or the one without their effect ($\tau = 1.0$ and $K = |\mathcal{D}|/2$).
\cref{tab:ablation1} indicates that maintaining a balanced $\tau$ to filter out overly similar images is crucial. 
Likewise, for a fixed $\tau$, a moderate $K$ further enhances the performance.
Thus, a proper combination of $\tau$ and $K$ is crucial to enhance performance to encourage a model to differentiate and adapt within various referring scenarios.
\begin{table}[h] 
\centering
\footnotesize
\caption{Cross-Effect on $\tau$ and $K$ (G-Ref UMD validation split)}
\label{tab:ablation1}
\newcolumntype{Y}{>{\centering\arraybackslash}X}
\begin{tabularx}{0.8\columnwidth}{@{}YY|YYYYY@{}}
    \toprule
    $\tau$ & $K$ & $P@0.5$ & $P@0.7$ & $P@0.9$ & oIoU & mIoU \\
    \midrule
    1.00 & 200 & 69.11 & 55.95 & 14.99 & 57.74 & 60.30 \\
    0.75 & 200 & \bf{70.01} & \bf{56.60} & \bf{15.89} & \bf{58.47} & \bf{61.36} \\
    1.00 & $\mathcal{D}|/2$ & 69.14 & 54.76 & 15.09 & 58.06 & 60.73 \\
    0.75 & $\mathcal{D}|/2$ & 69.00 & 57.09 & 16.06 & 58.15 & 60.90 \\
    \bottomrule
\end{tabularx}
\end{table}

%

\subsection{Further Discussion on Mosaic Design}

Unlike existing approaches~\cite{bochkovskiy2020yolov4}, our paper presents a $2 \times 2$ mosaic augmentation with a fixed cross-point.
One might argue that other tile patterns could be equally effective and question why the $2 \times 2$ configuration without cross point shift is deemed optimal.
Thus, to ascertain the efficacy of our $2 \times 2$ mosaic design with a fixed cross-point, we conduct a series of ablation studies on G-Ref.

\begin{figure}[h!]
    \centering
    \footnotesize
    \includegraphics[width=\linewidth]{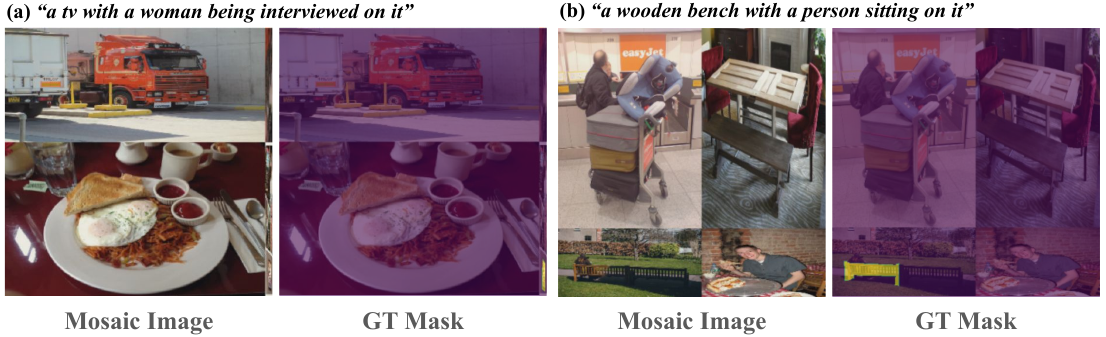} 
    \caption{Failure Cases of Cross-point shifted Mosaics.}
    \label{fig:randomcrs_failure}
\end{figure}

\vspace{0.1cm} \noindent
\textbf{Cross-point Shift.}
We test varying the cross-point's position, resizing images accordingly.
Specifically, we first arbitrarily move the cross-point, which sometimes placing it near the edge of the mosaic, making some objects hard to recognize.
We then constrain the movement within the central quarter of the image to ensure at least a quarter of the original image size.
However, these modifications do not improve performance, as indicated in \cref{tab:ablation2}, mainly due to the disrupted image ratio.
\cref{fig:randomcrs_failure}(b) illustrates that arbitrary cross-point movements can excessively compress one of the images, rendering smaller objects unrecognizable.

\vspace{0.1cm} \noindent
\textbf{Mosaic Grid Shape.}
We also compare the effectiveness of $2 \times 2$ and $3 \times 3$ grids in \cref{tab:ablation3}, with results showing that the $2 \times 2$ configuration is superior to the $3 \times 3$ setup.
This outcome suggests that maintaining a fixed cross point in a $2 \times 2$ layout preserves the semantic integrity of objects better than other configurations.
For aforementioned issues, we opt to maintain a fixed cross point in our final method.

\begin{table}[t]
\centering
\footnotesize
\caption{Ablation on Mosaic Configuration (G-Ref UMD validation split)}
\label{tab:ablation3}
\newcolumntype{Y}{>{\centering\arraybackslash}X}
\begin{tabularx}{0.8\columnwidth}{@{}lYYYYY@{}}
    \toprule
    & $P@0.5$ & $P@0.7$ & $P@0.9$ & oIoU & mIoU \\
    \midrule
    \multicolumn{1}{l}{$2 \times 2$} & 70.01 & \textbf{56.60} & \textbf{15.89} & \textbf{58.47} & \textbf{61.36} \\
    \multicolumn{1}{l}{$3 \times 3$} & \textbf{70.22} & 56.34 & 15.05 & 57.99 & 61.14\\
    \bottomrule
\end{tabularx}
\end{table}
\begin{table}[t]
\centering
\footnotesize
\caption{Ablation on Center Point Variation (G-Ref UMD validation split)}
\label{tab:ablation2}
\newcolumntype{Y}{>{\centering\arraybackslash}X}
\begin{tabularx}{\columnwidth}{@{}lYYYYY@{}}
    \toprule
    & $P@0.5$ & $P@0.7$ & $P@0.9$ & oIoU & mIoU \\
    \midrule
    Locating center point anywhere & 69.12 & 56.31 & 14.93 & 57.19 & 60.00 \\
    Locating within $H/4 \times W/4$ center block & 69.22 & 56.52 & 14.75 & 57.34 & 60.51\\
    Fixed center point in the middle & \bf{70.01} & \bf{56.60} & \bf{15.89} & \bf{58.47} & \bf{61.36} \\
    \bottomrule
\end{tabularx}
\end{table}
\begin{figure}[t]
    \centering
    \includegraphics[width=0.8\linewidth]{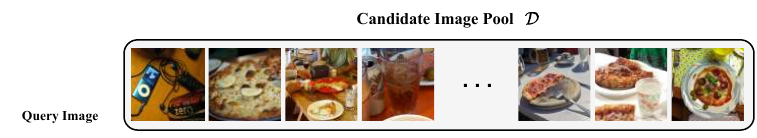}

    \includegraphics[width=0.8\linewidth]{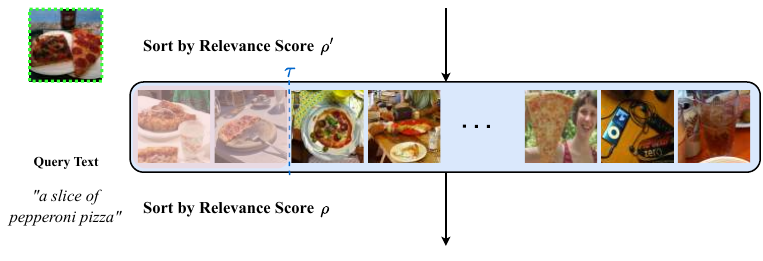}

    ~~~~~~~~~~~~~~~~
    \includegraphics[width=0.5\linewidth]{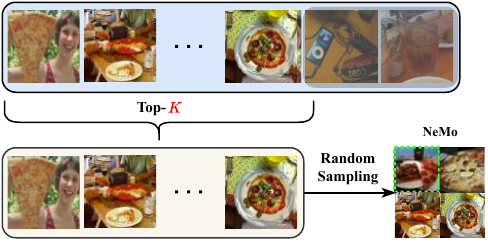}
    \caption{\textbf{Overall pipeline of NeMo when $\rho \neq \rho'$.} Given an image and a query, NeMo aims to select negative images at a proper level of difficulty. It filters out images that are too relevant and thus may cause false negatives by image-to-image similarity $\rho'$, and excludes irrelevant (easy) images by text-to-image retrieval $\rho$. Among the remaining candidates, it randomly selects 3 negative images to be included in the mosaic.}
    \label{fig:overall_arch2}
\end{figure}





\section{Similarity Score Options for Enhanced Filtering}
\label{sec:sim_score_opt}
To accurately eliminate false positives and negatives,
we evaluate the relevance $\rho$ between each candidate image $I^{(i)}$ and the referring expression $T$.
As detailed in \cref{sec:neg_img_selection}, setting an optimal upper bound is crucial for removing excessively relevant images to the $T$, thereby facilitating more effective training.
We propose using either text-to-image or image-to-image similarity as the relevance metric $\rho'$ to establish this upper bound, detailed below.

\vspace{0.1cm} \noindent
\textbf{Case 1: When $\rho = \rho'$.} 
This scenario corresponds to using text-to-image similarity to determine the upper bound.
We exclude images where $\rho_\text{t-i}^{(i)} \ge \tau_\text{t-i}$, following the simplest pipeline structure depicted in \cref{fig:overall_arch}.
This approach requires only a single sorting process for the entire pipeline for both upper and lower bounds, since it relies solely on a single metric, the text-to-image relevance score.

\vspace{0.1cm} \noindent
\textbf{Case 2: When $\rho \neq \rho'$.}
In this case, we use different metric to determine the upper and lower bound; for example, we use image-to-image similarity to determine the upper bound, filtering out images satisfying $\rho_\text{i-i}^{(i)} \ge \tau_\text{i-i}$, while keep text-to-image similarity for the lower bound.
With $\rho \neq \rho'$, sorting needs to be conducted twice.
Initially, we sort images based on their image-to-image relevance to eliminate overly similar images to the referring text.
Following this exclusion, the remaining images are sorted according to their text-to-image scores to establish the lower bound.
\cref{fig:overall_arch2} illustrates this procedure.


\section{Determining the Lower-bound and Upper-bound}
\label{sec:lower_bound}

In this section, we provide empirical results for scenarios where $\rho = \rho'$ and $\rho \neq \rho'$, proposed in \cref{sec:neg_img_selection}.
We conduct this study using the CRIS~\cite{wang2022cris} model on the UMD validation set of G-Ref.
\begin{table}[h]
\centering
\caption{Determining $K$ on Different Relevance Metric}
\label{tab:compare_simscore_K}
\footnotesize
\renewcommand{\tabcolsep}{6pt}
\resizebox{0.82\linewidth}{!}{
  \begin{tabular}{l|ccccc|ccccc}
    \toprule 
    \multirow{2}{*}{$K$} & \multicolumn{5}{c|}{$\rho = \rho'$. $\tau=0.75$} & \multicolumn{5}{c}{$\rho \neq \rho'$. $\tau=0.25$} \\
    & P@0.5 & P@0.7 & P@0.9 & oIoU & mIoU & P@0.5 & P@0.7 & P@0.9 & oIoU & mIoU \\
    \midrule
    10 & 69.06 & 56.62 & 15.48 & 57.02 & 60.06 & 69.46 & 56.03 & 15.79 & 57.26 & 60.56 \\
    50 & 68.89 & 54.86 & 15.24 & 57.29 & 60.26 & 70.03 & 56.42 & 14.38 & 57.58 & 60.83 \\
    100 & 68.69 & 54.94 & 14.13 & 57.13 & 59.85 & 69.57 & 56.52 & 14.20 & 57.93 & 60.54 \\
    200 & 68.61 & 55.02 & 14.85 & 57.54 & 60.39 & 70.01 & 56.60 & 15.89 & 58.47 & 61.36 \\
    300 & 69.72 & 56.83 & 15.87 & 58.59 & 61.02 & 70.02 & 55.98 & 14.48 & 57.85 & 61.00 \\ 
    500 & 69.83 & 56.86 & 15.24 & 58.36 & 60.97 & 68.95 & 55.80 & 15.36 & 57.24 & 60.49 \\
    \bottomrule
  \end{tabular}}
\end{table}

\begin{table}[h]
    \caption{Ablation study on $\tau$}
    \label{tab:suppl_lower_bound}
    \begin{subtable}[h]{0.49\textwidth}
        \centering
        \caption{$\rho = \rho', K = 300$}
        \label{tab:week1}
        \renewcommand{\tabcolsep}{5pt}
        \resizebox{0.99\linewidth}{!}{
        \begin{tabular}{l|ccccc}
        \toprule
        $\tau$ & P@0.5 & P@0.7 & P@0.9 & oIoU & mIoU \\
        \midrule
        0.40 & 68.10 & 55.39 & 14.67 & 57.00 & 59.69  \\
        0.35 &  68.69 & 55.83 & 14.37 & 56.77 & 60.40 \\
        0.30 & 69.25 & 56.27 & 15.16 & 57.47 & 60.89  \\
        0.25 & \textbf{69.72} & \textbf{56.83} & \textbf{15.87} & \textbf{58.59} & \textbf{61.02} \\
        0.20 & 68.12 & 54.97 & 14.87 & 56.90 & 59.38 \\
        \bottomrule
        \end{tabular}}
    \end{subtable}
    \hfill
    \begin{subtable}[h]{0.49\textwidth}
        \centering
     \caption{$\rho \neq \rho', K = 200 $}
     \label{tab:temps}
     \renewcommand{\tabcolsep}{5pt}
        \resizebox{0.99\linewidth}{!}{
        \begin{tabular}{l|ccccc}
        \toprule
        $\tau$ & P@0.5 & P@0.7 & P@0.9 & oIoU & mIoU \\
        \midrule
        1.00 & 69.14 & 55.76 & 15.09 & 58.06 & 60.73 \\
        0.85 &  69.47 &  56.29 &  15.08 &  57.57 & 60.50 \\
        0.75 &  \textbf{70.01} &  \textbf{56.60} & \textbf{15.89} &  \textbf{58.47} & \textbf{61.36} \\
        0.60 &  69.63 & 56.17 & 14.07 & 57.92 & 60.36  \\
        \bottomrule
        \end{tabular}}
    \end{subtable}
\end{table}

\vspace{0.1cm} \noindent
\textbf{Negative Image Pool Size $K$.}
We determine the initial value for $\tau$ by analyzing the relevance metric's distribution, used for setting the upper bounds.
Scores are normalized, and the cutoff is established at the top 5\%, corresponding to 0.75 when $\rho \neq \rho'$ and 0.25 when $\rho = \rho'$.
\cref{tab:compare_simscore_K} indicates that both excessively low and high values of $K$ lead to suboptimal performance.
The best performance is achieved at $K=200$ when $\rho \neq \rho'$ and at $K=300$ for $\rho = \rho'$.

\vspace{0.1cm} \noindent
\textbf{Upper-bound Threshold $\tau$.}
Fixing the $K$ to the optimal value found above, we explore the optimal $\tau$.
When $\rho = \rho'$, we examine potential $\tau$ values from $\{0.2, 0.25, 0.3, 0.35, 0.44\}$.
Here, the maximum $\tau = 0.44$ with the text-to-image similarity implies no upper bound.
Conversely, in cases where $\rho \neq \rho'$, we assessed $\tau$ values from $\{0.6, 0.75, 0.85, 1.0\}$. Selecting a maximum $\tau$ value of 1 for image-to-image similarity comparisons effectively removes any limitation on filtering images closely resembling the target, similar to the previous scenario.
As discovered in \cref{tab:suppl_lower_bound}, a larger $\tau$ results in fewer images being filtered out, increasing the likelihood of false positives/negatives.
Conversely, too small a $\tau$ value approximates random sampling, undermining the efficiency of the model.

\vspace{0.1cm} \noindent
\textbf{Overall Comparison.} Irrespective of the relevance metric used to define upper bounds, the peak performance in terms of mIoU and oIoU is seen at $K=300$ and $\tau=0.25$ for $\rho = \rho'$, and at $K=200$ and $\tau=0.75$ for $\rho \neq \rho'$. Notably, using image-to-image similarity scores ($\rho \neq \rho'$) slightly enhances outcomes.

\section{Experiments on False Positives}
\label{sec:fp_exp}

Based on our observation that False Positives (FP) arise when positional keywords present in the text query but are overlooked in mosaic creation, we design an experiment to mitigate FP occurrences.
When positional keywords, \emph{e.g.}, those in~\cref{tab:fp_exp_pos}, are present, we arrange each target image in the mosaic corresponding only to the allowed locations.

\begin{figure}
    \centering
    \includegraphics[width=0.6\linewidth]{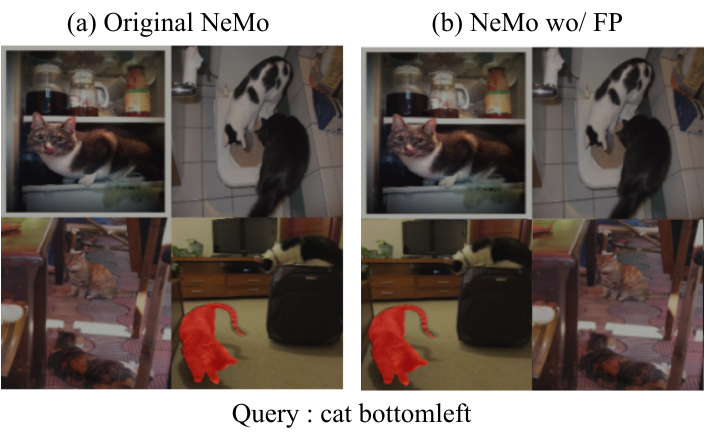}
    \captionof{figure}{An example of FP experiment. The ground truth segmentation map is highlighted in red.} 
    \label{fig:fp_exp_sample}
\end{figure}

\cref{fig:fp_exp_sample} illustrates an example, where an image tagged with `cat bottomleft'.
The cat in this image original image becomes invalid after augmentation in \cref{fig:fp_exp_sample}(a), as it is located on the right side of the augmented image and there exists another cat on the bottom left.
To prevent such a case, we restrict the position to be located when the referring expression contains a positioning keyword listed in \cref{tab:fp_exp_pos}, resulting in a valid augmentation shown in \cref{fig:fp_exp_sample}(b).
We reduce false positives by applying these positional restrictions, and assess the impact on performance compared to the standard NeMo framework.
Interestingly, \cref{tab:fp_exp} reports the difference between these two is insignificant, indicating that adjusting hyperparameters $\tau$ and $K$ effectively mitigates the FP issue.

\begin{table}
    \centering
    \caption{Each possible coordinate on the mosaic image represents a specific corner.}
    \label{tab:fp_exp_pos}
    \begin{tabular}{c|c}
    \toprule
        positional & allowed \\
        keywords & location \\
        \midrule
        top, high, above & upper-left, upper-right \\
        left & upper-left, lower-left \\
        right & upper-right, lower-right \\
        bottom, low, below & lower-left, lower-right \\ 
        \bottomrule
    \end{tabular}
\end{table}

\end{document}